\documentclass[10pt,twocolumn,letterpaper]{article}

\usepackage[pagenumbers]{cvpr} 

\usepackage{graphicx}
\usepackage{amsmath}
\usepackage{amssymb}
\usepackage{booktabs, multirow}
\usepackage[accsupp]{axessibility}

\usepackage[pagebackref,breaklinks,colorlinks]{hyperref}

\usepackage[capitalize]{cleveref}
\crefname{section}{Sec.}{Secs.}
\Crefname{section}{Section}{Sections}
\Crefname{table}{Table}{Tables}
\crefname{table}{Tab.}{Tabs.}

\def\cvprPaperID{*****} 
\def\confName{CVPR}
\def\confYear{2022}
\newcommand\blfootnote[1]{%
  \begingroup
  \renewcommand\thefootnote{}\footnote{#1}%
  \addtocounter{footnote}{-1}%
  \endgroup
}

\begin{document}

\title{NTIRE 2022 Challenge on Super-Resolution and Quality Enhancement of Compressed Video: Dataset, Methods and Results}

\author
{Ren Yang\and Radu Timofte \and 
Meisong Zheng\and Qunliang Xing\and Minglang Qiao\and Mai Xu\and Lai Jiang\and Huaida Liu\and Ying Chen\and
Youcheng Ben\and Xiao Zhou\and Chen Fu\and Pei Cheng\and Gang Yu\and
Junyi Li\and Renlong Wu\and Zhilu Zhang\and Wei Shang\and Zhengyao Lv\and Yunjin Chen\and Mingcai Zhou\and Dongwei Ren\and Kai Zhang\and Wangmeng Zuo\and
Pavel Ostyakov\and Vyal Dmitry\and Shakarim Soltanayev\and Chervontsev Sergey\and Zhussip Magauiya\and Xueyi Zou\and Youliang Yan\and 
Pablo Navarrete Michelini\and Yunhua Lu\and 
Diankai Zhang\and Shaoli Liu\and Si Gao\and Biao Wu\and Chengjian Zheng\and Xiaofeng Zhang\and Kaidi Lu\and Ning Wang\and
Thuong Nguyen Canh\and Thong Bach\and
Qing Wang\and Xiaopeng Sun\and Haoyu Ma\and Shijie Zhao\and Junlin Li\and
Liangbin Xie\and Shuwei Shi\and Yujiu Yang\and Xintao Wang\and Jinjin Gu\and Chao Dong\and
Xiaodi Shi\and Chunmei Nian\and Dong Jiang\and Jucai Lin\and
Zhihuai Xie\and
Mao Ye\and Dengyan Luo\and Liuhan Peng\and Shengjie Chen\and
Xin Liu\and 
Qian Wang\and  Xin Liu\and
Boyang Liang\and Hang Dong\and Yuhao Huang\and Kai Chen\and
Xingbei Guo\and Yujing Sun\and Huilei Wu\and Pengxu Wei\and
Yulin Huang\and Junying Chen\and
Ik Hyun Lee\and Sunder Ali Khowaja\and Jiseok Yoon
}
\maketitle

\begin{abstract}
   This paper reviews the NTIRE 2022 Challenge on Super-Resolution and Quality Enhancement of Compressed Video. In this challenge, we proposed the LDV 2.0 dataset, which includes the LDV dataset (240 videos) and 95 additional videos. This challenge includes three tracks. Track 1 aims at enhancing the videos compressed by HEVC at a fixed QP. Track 2 and Track 3 target both the super-resolution and quality enhancement of HEVC compressed video. They require x2 and x4 super-resolution, respectively. The three tracks totally attract more than 600 registrations. In the test phase, 8 teams, 8 teams and 12 teams submitted the final results to Tracks 1, 2 and 3, respectively. The proposed methods and solutions gauge the state-of-the-art of super-resolution and quality enhancement of compressed video. The proposed LDV 2.0 dataset is available at \url{https://github.com/RenYang-home/LDV_dataset}.
   The homepage of this challenge (including open-sourced codes) is at \url{https://github.com/RenYang-home/NTIRE22_VEnh_SR}.
\end{abstract}

\section{Introduction}
\blfootnote{Ren Yang ({\tt ren.yang@vision.ee.ethz.ch}, ETH Z\"urich) and Radu Timofte ({\tt radu.timofte@uni-wuerzburg.de}, ETH Z\"urich, Julius Maximilian University of W\"urzburg) are the organizers of the NTIRE 2022 challenge, and other authors are the participants.\\
The Appendix lists the authors’ teams and affiliations.\\
NTIRE 2022 website: \url{https://data.vision.ee.ethz.ch/cvl/ntire22/}}

Nowadays, there are increasing demands on transmitting high quality videos over the Internet. Video compression plays an important role on the efficient video transmission through the band-limited Internet, however, it also unavoidably lead to compression artifacts, which may severely degrade the visual quality. Therefore, it is necessary to study on the quality enhancement of compressed video (Track 1). Besides, in early years, due to the much lower speed of the Internet and the smaller memory of devices, the videos were usually with low resolution. Therefore, in the case that we intend to restore these videos to high resolution and better quality, it is meaningful to explore the methods that enhance the quality and meanwhile achieve the super-resolution of compressed video (Track 2 and Track 3).

For enhancing the quality of compressed video, there has been plenty of works proposed in the past a few years~\cite{yang2017decoder, yang2018enhancing, wang2017novel, lu2018deep, yang2018multi, xu2021multi, yang2019quality, guan2019mfqe, Xu_2019_ICCV, deng2020spatio, yang2020learning, huo2021recurrent, wang2020multi}. In these methods, \cite{yang2017decoder, yang2018enhancing, wang2017novel} are single-frame quality enhancement methods, while~\cite{lu2018deep, yang2018multi, yang2019quality, guan2019mfqe, Xu_2019_ICCV, deng2020spatio, yang2020learning, huo2021recurrent, wang2020multi} propose enhancing quality by taking advantage of temporal correlation. Besides, a great number of methods~\cite{tao2017detail,caballero2017real,liu2017robust,sajjadi2018frame,tian2020tdan,wang2019edvr,huang2015bidirectional,jo2018deep,haris2019recurrent,wang2020deep} were proposed for video super-resolution. However, these methods only focus on the super-resolution of uncompressed videos. In 2020, Chen~\etal~\cite{chen2020bitstream} proposed the compressed domain deep video super-resolution method, and their latest work~\cite{chen2021compressed} further improves the quality performance.

The NTIRE 2022 Challenge on Super-Resolution and Quality Enhancement of Compressed Video is a step forward for establishing a benchmark of video quality enhancement and a benchmark of super-resolution on compressed video. It uses the LDV 2.0 dataset, which contains 335 videos with the diversities of content, motion, frame-rate, \etc. In the following, we first describe the NTIRE 2022 Challenge, including the proposed LDV 2.0 dataset. Then, we will introduce the proposed methods and the results.

\section{NTIRE 2022 Challenge}

The objectives of the NTIRE 2022 challenge on Super-Resolution and Quality Enhancement of Compressed Video are: (i) to advance the state-of-the-art in quality enhancement of compressed video; (i) to advance the state-of-the-art in super-resolution of compressed video; (iii) to compare different solutions; (iv) to promote the LDV 2.0 dataset.

This challenge is one of the NTIRE 2022 associated challenges: spectral recovery~\cite{arad2022ntirerecovery}, spectral demosaicing~\cite{arad2022ntiredemosaicing},
perceptual image quality assessment~\cite{gu2022ntire},
inpainting~\cite{romero2022ntire},
night photography rendering~\cite{ershov2022ntire},
efficient super-resolution~\cite{li2022ntire},
learning the super-resolution space~\cite{lugmayr2022ntire},
super-resolution and quality enhancement of compressed video~\cite{yang2022ntire},
high dynamic range~\cite{perezpellitero2022ntire},
stereo super-resolution~\cite{wang2022ntire},
burst super-resolution~\cite{bhat2022ntire}.

\subsection{LDV 2.0 dataset}

The proposed LDV 2.0 dataset is an enlarged version of the LDV dataset~\cite{yang2021dataset} with 95 additional videos. Therefore, there are totally 335 videos in the LDV 2.0 dataset. The same as LDV, the additional videos in LDV 2.0 are collected from YouTube~\cite{youtube}, containing 10 categories of scenes, \ie, \textit{animal}, \textit{city}, \textit{close-up}, \textit{fashion}, \textit{human}, \textit{indoor}, \textit{park}, \textit{scenery}, \textit{sports} and \textit{vehicle}, and they are with diverse frame-rates from 24 fps to 60 fps. To ensure the high quality of the groundtruth videos, we only collect the videos with 4K resolution, and without obvious compression artifacts. We downscale the videos by the factor of $4$ using the Lanczos filter~\cite{turkowski1990filters} to further remove the artifacts, and crop the width and height of each video to the multiples of 8, due to the requirement of the HEVC test model (HM). Besides, we convert videos to the format of YUV 4:2:0, which is the most commonly used format in the existing literature. Note that all source videos in our LDV 2.0 dataset have the licence of \emph{Creative Commons Attribution licence (reuse allowed)}\footnote{\url{https://support.google.com/youtube/answer/2797468?hl=en}}, and our LDV 2.0 dataset is used for academic and research proposes.

In the NTIRE 2022 Challenge, we use the 240 videos in the original LDV dataset~\cite{yang2021dataset} as the training sets for all three tracks. We choose 90 videos from the remaining videos in LDV 2.0, and split them into six datasets with 15 videos in each. When splitting the datasets, we have paid attention to the diversity (content, frame rate, \etc) of the videos in each set. These six datasets are utilized as the validation and test sets for the three tracks, respectively. All videos in the LDV and LDV 2.0 datasets and the splits in NTIRE 2021 and NTIRE 2022 Challenges are publicly available at \url{https://github.com/RenYang-home/LDV_dataset}.

\begin{table*}[!t]
\scriptsize
  \centering
  \caption{Results of Track 1 (quality enhancement). \textcolor{blue}{Blue} indicates the state-of-the-art method.}
  \vspace{-1em}
    \begin{tabular}{ccccccccccccccccc}
    \cmidrule[\heavyrulewidth]{1-17}
    \multirow{2}[3]{*}{Team} & \multicolumn{16}{c}{PSNR (dB)}  \\
\cmidrule{2-17}        & \#1 & \#2 & \#3 & \#4 & \#5 & \#6 & \#7 & \#8 & \#9 & \#10 & \#11 & \#12 & \#13 & \#14 & \#15 & Ave. \\
\cmidrule{1-17}    
TaoMC2 & 34.20  & 32.71  & 31.59  & 37.12  & \textbf{32.75}  & 30.81  & \textbf{30.79}  & \textbf{34.58}  & \textbf{32.13}  & \textbf{37.13}  & \textbf{27.35}  & \textbf{24.42}  & \textbf{30.60}  & \textbf{33.38}  & 31.54  & \textbf{32.07}  \\ 
        GY-Lab & \textbf{34.23}  & \textbf{32.90}  & \textbf{31.61}  & \textbf{37.32}  & 32.70  & \textbf{30.83}  & 30.77  & 34.55  & 32.06  & 36.87  & 27.34  & 24.28  & 30.58  & 33.34  & \textbf{31.56}  & 32.06   \\ 
        HIT\&ACE & 34.07  & 32.68  & 31.47  & 37.10  & 32.60  & 30.73  & 30.70  & 34.38  & 31.98  & 36.69  & 27.27  & 24.38  & 30.54  & 33.20  & 31.40  & 31.94   \\ 
        BOE-IOT-AIBD & 33.96  & 32.58  & 31.45  & 36.90  & 32.45  & 30.63  & 30.54  & 34.25  & 31.91  & 36.51  & 27.16  & 24.05  & 30.47  & 33.12  & 31.33  & 31.82   \\ 
        OCL-VCE & 33.81  & 32.54  & 31.30  & 36.82  & 32.28  & 30.59  & 30.47  & 34.07  & 31.75  & 36.28  & 27.08  & 24.00  & 30.37  & 33.01  & 31.25  & 31.71   \\ 
        \textcolor{blue}{BasicVSR++~\cite{chan2021basicvsr++}} & 33.73  & 32.42  & 31.22  & 36.75  & 32.16  & 30.57  & 30.41  & 33.99  & 31.68  & 36.20  & 27.06  & 23.93  & 30.24  & 32.94  & 31.20  & 31.63   \\ 
        OREO & 33.64  & 32.38  & 31.16  & 36.80  & 32.08  & 30.55  & 30.37  & 34.02  & 31.63  & 36.12  & 27.02  & 23.98  & 30.30  & 32.91  & 31.10  & 31.60   \\ 
        UESTC+XJU CV & 33.57  & 32.33  & 31.04  & 36.60  & 31.97  & 30.46  & 30.24  & 33.76  & 31.47  & 35.96  & 26.94  & 23.77  & 30.19  & 32.77  & 31.05  & 31.47   \\ 
        AVRT & 33.19  & 31.19  & 30.14  & 35.59  & 31.37  & 29.87  & 29.88  & 33.15  & 31.05  & 35.79  & 26.81  & 23.61  & 29.39  & 32.32  & 29.77  & 30.88   \\ 
    \midrule
    
    Compressed video & 32.43  & 30.18*  & 29.05  & 34.31  & 30.49  & 28.99  & 28.84  & 31.80  & 29.86  & 34.65  & 26.30  & 22.87  & 28.42  & 31.21  & 29.03  & 29.90 \\
    
    \cmidrule[\heavyrulewidth]{1-17}
     \multicolumn{16}{l}{* The frames with MSE = 0 are excluded when calculating the average PSNR.}
    \end{tabular}%
  \label{tab:track1}%
\end{table*}%

\begin{table*}[!t]
\scriptsize
  \centering
  \caption{Results of Track 2 (quality enhancement and $\times 2$ super-resolution)}
  \vspace{-1em}
    \begin{tabular}{ccccccccccccccccc}
    \cmidrule[\heavyrulewidth]{1-17}
    \multirow{2}[3]{*}{Team} & \multicolumn{16}{c}{PSNR (dB)}  \\
\cmidrule{2-17}        & \#1 & \#2 & \#3 & \#4 & \#5 & \#6 & \#7 & \#8 & \#9 & \#10 & \#11 & \#12 & \#13 & \#14 & \#15 & Ave. \\
\cmidrule{1-17}   

TaoMC2 & \textbf{27.71}  & \textbf{24.11}  & \textbf{26.53}  & \textbf{31.30}  & \textbf{31.84}  & \textbf{27.11}  & \textbf{25.13}  & \textbf{24.20}  & \textbf{28.76}  & \textbf{32.16}  & \textbf{26.39}  & \textbf{27.83}  & \textbf{26.27}  & \textbf{24.73}  & \textbf{29.17}  & \textbf{27.55}  \\ 
        GY-Lab & 27.56  & 23.89  & 26.47  & 31.26  & 31.69  & 27.03  & 25.05  & 24.12  & 28.61  & 32.15  & 26.34  & 27.71  & 26.08  & 24.61  & \textbf{29.17}  & 27.45   \\ 
        HIT\&ACE & 27.43  & 23.79  & 26.38  & 31.13  & 31.42  & 26.96  & 24.94  & 23.91  & 28.38  & 32.01  & 26.21  & 27.61  & 25.95  & 24.47  & 29.06  & 27.31   \\ 
        ZX\_VIP & 27.46  & 23.77  & 26.42  & 31.03  & 31.50  & 26.83  & 24.91  & 24.00  & 28.46  & 31.69  & 26.23  & 27.61  & 25.97  & 24.55  & 29.00  & 27.30   \\ 
        Trick collector & 27.44  & 23.90  & 26.40  & 31.02  & 31.38  & 26.94  & 25.00  & 22.28  & 28.34  & 31.86  & 26.17  & 27.58  & 26.03  & 24.49  & 29.01  & 27.19   \\ 
        HMSR & 27.18  & 23.54  & 26.26  & 30.85  & 31.22  & 26.69  & 24.68  & 23.84  & 28.28  & 31.61  & 26.13  & 27.45  & 25.67  & 24.34  & 28.68  & 27.09   \\ 
        TBE & 26.87  & 23.23  & 26.12  & 30.53  & 30.76  & 26.36  & 24.36  & 23.65  & 27.92  & 31.27  & 25.93  & 27.23  & 25.28  & 24.07  & 28.34  & 26.80   \\ 
        AVRT & 26.58  & 22.88  & 25.92  & 30.21  & 30.39  & 26.23  & 24.01  & 23.43  & 27.68  & 30.77  & 25.77  & 26.94  & 24.70  & 23.68  & 27.95  & 26.48   \\ 
    \midrule
    Bicubic $\times 2$ & 25.55 &	22.03 &	25.53 &	29.12 &	29.33 &	25.20 &	23.17 &	22.67 &	26.54 &	29.65 &	25.12 &	26.25 &	23.92 &	22.76 &	26.83 &	25.58 \\

    \cmidrule[\heavyrulewidth]{1-17}
    \end{tabular}%
  \label{tab:track2}%
\end{table*}%

\begin{table*}[!t]
\scriptsize
  \centering
  \caption{Results of Track 3 (quality enhancement and $\times 4$ super-resolution). \textcolor{blue}{Blue} indicates the state-of-the-art method.}
  \vspace{-1em}
    \begin{tabular}{ccccccccccccccccc}
    \cmidrule[\heavyrulewidth]{1-17}
    \multirow{2}[3]{*}{Team} & \multicolumn{16}{c}{PSNR (dB)}  \\
\cmidrule{2-17}        & \#1 & \#2 & \#3 & \#4 & \#5 & \#6 & \#7 & \#8 & \#9 & \#10 & \#11 & \#12 & \#13 & \#14 & \#15 & Ave. \\
\cmidrule{1-17}    
GY-Lab & 27.13  & 21.22  & \textbf{25.15}  & \textbf{24.78}  & \textbf{26.80}  & 24.39  & 21.93  & \textbf{24.32}  & \textbf{28.05}  & 22.33  & 22.24  & \textbf{24.06}  & 20.88  & \textbf{25.66}  & 24.49  & \textbf{24.23}   \\ 
        TaoMC2 & 27.20  & \textbf{21.24}  & 25.14  & 24.71  & 26.72  & 24.19  & 21.98  & 24.31  & 28.04  & \textbf{22.35}  & 22.24  & \textbf{24.06}  & 20.93  & 25.63  & 24.49  & 24.22   \\ 
        NoahTerminalCV & \textbf{28.40}  & 20.84  & 24.61  & 23.91  & 26.44  & \textbf{25.66}  & \textbf{22.06}  & 23.83  & 26.79  & 21.86  & \textbf{23.81}  & 23.75  & \textbf{21.56}  & 24.90  & \textbf{24.61} & 24.20   \\ 
        HIT\&ACE & 27.06  & 21.10  & 24.99  & 24.49  & 26.33  & 23.93  & 21.83  & 24.18  & 27.74  & 22.18  & 22.13  & 23.95  & 20.75  & 25.48  & 24.30  & 24.03   \\ 
        XPixel & 27.02  & 21.10  & 24.96  & 24.58  & 26.43  & 24.04  & 21.80  & 24.15  & 27.48  & 22.15  & 22.14  & 23.98  & 20.81  & 25.42  & 24.26  & 24.02   \\ 
        Trick collector & 26.75  & 21.15  & 25.01  & 24.01  & 26.23  & 23.85  & 21.04  & 24.16  & 27.66  & 22.12  & 22.10  & 23.88  & 20.69  & 25.38  & 24.26  & 23.88   \\ 
        HyperPixel & 26.72  & 20.87  & 24.68  & 23.82  & 25.92  & 23.49  & 21.46  & 23.90  & 27.16  & 21.90  & 22.04  & 23.66  & 20.51  & 25.12  & 23.88  & 23.68   \\ 
        CVStars & 26.75  & 20.88  & 24.72  & 23.66  & 25.90  & 23.71  & 21.52  & 23.81  & 26.94  & 21.86  & 22.16  & 23.77  & 20.50  & 25.00  & 23.90  & 23.67   \\ 
        AVRT & 26.63  & 20.75  & 24.39  & 23.58  & 25.72  & 23.48  & 21.39  & 23.69  & 26.64  & 21.76  & 22.07  & 23.54  & 20.41  & 24.98  & 23.70  & 23.52   \\ 
        StarRay & 26.46  & 20.66  & 24.17  & 22.97  & 24.15  & 23.22  & 21.19  & 23.37  & 26.01  & 21.49  & 22.10  & 23.41  & 20.22  & 24.39  & 23.48  & 23.15   \\ 
        \textcolor{blue}{CDVSR~\cite{chen2021compressed}}* & 26.20  & 20.64  & 24.03  & 22.61  & 24.90  & 23.02  & 21.02  & 23.24  & 25.88  & 21.35  & 21.97  & 23.18  & 20.13  & 23.89  & 23.40  & 23.03   \\ 
        Modern\_SR & 26.38  & 20.49  & 23.90  & 22.76  & 24.81  & 22.93  & 21.09  & 23.13  & 25.56  & 21.29  & 21.92  & 23.27  & 20.14  & 24.18  & 23.31  & 23.01   \\ 
        TUK-IKLAB & 26.19  & 20.55  & 23.74  & 22.37  & 24.17  & 22.39  & 20.80  & 22.84  & 25.10  & 21.07  & 21.94  & 23.05  & 19.89  & 23.35  & 23.12  & 22.70   \\ 
    
    \midrule
    Bicubic $\times 4$ & 26.04 &	20.52 &	23.53 &	22.14 &	24.24 &	22.06 &	20.58 &	22.63 &	24.80 &	20.89 &	21.91 &	22.92 &	19.78 &	23.32 &	23.06 &	22.56 \\

    \cmidrule[\heavyrulewidth]{1-17}
    
    \multicolumn{16}{l}{* The CDVSR~\cite{chen2021compressed} method only enhances the Y channel and upsamples the U and V channels by the bicubic algorithm.}
    \end{tabular}%
  \label{tab:track3}%
\end{table*}%

\begin{table*}[!t]
\scriptsize
  \centering
  \caption{The time complexity, hardware, test strategies and training data of the proposed methods (reported by the participants).}
  \vspace{-1em}
    \begin{tabular}{ccccccc}
    \toprule
    \multirow{2}[4]{*}{Team} & \multicolumn{3}{c}{Running time (s) per frame} & \multirow{2}[4]{*}{Hardware} & \multirow{2}[4]{*}{Ensemble / Fusion} & \multirow{2}[4]{*}{Extra training data}\\
\cmidrule{2-4}          & Track 1 & Track 2 & Track 3 &       &       &  \\
    \midrule
TaoMC2 & 44.1 & 44.1 & 13.0&  Tesla V100 & Flip/rotation x8, two models& 870 videos from YouTube~\cite{youtube}\\
GY-Lab & 6.9 & 4.6 & 11.5 & Tesla V100 & Spatial-temporal ensemble and several models & REDS~\cite{nah2019ntire}, Vimeo90K~\cite{xue2019video}, YouTube~\cite{youtube}\\
HIT\&ACE & 17.30 & 10.97 & 17.40 &Tesla V100 &  Flip/rotation x8, two models &  540 samples from YouTube~\cite{youtube}\\
NoahTerminalCV & - & - & 150 &  Tesla V100&  Flip/rotation  x8, five networks & 90,000 videos from YouTube~\cite{youtube}\\
BOE-IOT-AIBD & 1.61 & - & - &  Tesla V100 & Flip/rotation  x8 & -\\
ZX\_VIP & - & 12 & - & Tesla V100& Flip/rotation  x8 & REDS \\
OCL-VCE & 28.72 & - & - &  Tesla T4 & Flip/rotation  x8 & -\\
Trick collector & - & 2.56 & 3.2 & Tesla A100 & Flip/rotation  x6/x8, model voting& REDS~\cite{nah2019ntire}\\
XPixel & - & - & 13.02 &  Tesla A100 & Flip/rotation  x8& REDS~\cite{nah2019ntire}, Vimeo90K~\cite{xue2019video} and 2174 clips\\
OREO& 19.4 &  &  & Tesla A40 & Flip/rotation  x8& -\\
HMSR &  & 14.36 &  &  Tesla A100 & Flip/rotation  x8& 1274 additional from Youtube~\cite{youtube}\\
UESTC+XJU CV & 0.16 & - & - &  GeForce RTX 3090  & - & - \\
TBE & - & 0.90 & - & Tesla V100& - & 91 videos\\
HyperPixel & - & - & 0.44 & Tesla V100 & Flip/rotation  x8 & -\\
CVStars & - & - & 10 & Tesla V100 &  Flip/rotation  x8, epoch-ensemble & - \\
AVRT & 27 & 8 & 2 & Tesla A100 & Flip/rotation  x4 & 202 videos\\
StarRay & - & - & 4.0 &  GeForce RTX 2080 Ti & Two models with different loss& -\\
Modern\_SR& - & - & 0.86 & GeForce RTX 3080 & - & -\\
TUK-IKLAB & - & - & $\leq$1.0 & GeForce RTX 3090 & - & -\\

    \bottomrule
    \end{tabular}%
  \label{tab:time}%
\end{table*}%

\subsection{Track 1 -- Quality enhancement}

To establish a progressive benchmark upon NTIRE 2021~\cite{yang2021ntire}, the Track 1 in NTIRE 2022 is set as the same task as that in NTIRE 2021. That is, Track 1 aims at enhancing the quality of compressed video towards fidelity. We evaluate the enhanced quality in terms of PSNR. In this track, the videos are compressed using the official HEVC test model (HM 16.20\footnote{\url{https://hevc.hhi.fraunhofer.de/svn/svn_HEVCSoftware/tags/HM-16.20}}) at QP = 37 the default Low-Delay P (LDP) setting (\textit{encoder\_lowdelay\_P\_main.cfg}).

\subsection{Track 2 -- Quality enhancement with x2 SR}

Track 2 is a more challenging task, which requires the participants to enhance and meanwhile $\times 2$ super-resolve the compressed video. In this track, the input videos are first downsampled by the following command:

\ 

\noindent\texttt{ffmpeg -pix\_fmt yuv420p -s WxH -i x.yuv \\ -vf scale=(W/2)x(H/2):flags=bicubic x\_down2.yuv}

\ 

\noindent where \texttt{x}, \texttt{W} and \texttt{H} indicates the video name, width and height, respectively. Then, the downsampled video is compressed by HM 16.20 with the same configurations as that in Track~1. Note that in this track, we first crop the groundtruth videos to make sure that the downsampled width (\texttt{W/2}) and height (\texttt{H/2}) are integer numbers.

\subsection{Track 3 -- Quality enhancement with x4 SR}

In Track 3, we further use $\times 4$ downsampling, and therefore the participants are required to enhance and meanwhile $\times 4$ super-resolve the compressed video. In this track, we change the width and height of downsampled video to \texttt{W/4} and \texttt{H/4}. Other settings are the same as Track 2. We also crop the groundtruth videos to make sure that the downsampled width (\texttt{W/4}) and height (\texttt{H/4}) are integer numbers.

\section{Challenge results}

\subsection{Track 1: Quality enhancement}

Table~\ref{tab:track1} shows the PSNR results of the 8 methods proposed in this challenge, in comparison with the compressed videos without enhancement. Besides, we also show the result of the winner method in NTIRE 2021 (BasicVSR++~\cite{chan2021basicvsr++}), which is the state-of-the-art performance.

It can be seen from Table~\ref{tab:track1} that the PSNR improvement of the 8 proposed methods ranges from 0.98 dB to 2.17 dB. The top 3 methods have the PSNR improvement higher than 2.0 dB, and the top 5 methods successfully outperform the winner method in NTIRE 2021 (BasicVSR++~\cite{chan2021basicvsr++}). Specifically, the TaoMC2 Team achieves the best average performance in this track, and its results are the best on 9 videos. The GY-Lab Team ranks second with very similar average performance to the TaoMC2 Team, and it has the best results on 6 out of the 15 test videos. The PSNR results of both these two top methods are higher than the last winner (BasicVSR++~\cite{chan2021basicvsr++}) by more than 0.4 dB.

\subsection{Track 2 -- Quality enhancement with x2 SR}

The results of Track 2 are tabulated in Table~\ref{tab:track2}. We also reported the PSNR of the videos that are directly upscaled by the bicubic algorithm, which can considered as the unprocessed videos. It can be see that the best method proposed by the TaoMC2 Team improves the PSNR by around 2.0 dB, in comparison with bicubic $\times 2$, and it also achieves the highest PSNR on all test videos. The GY-Lab Team is 0.1 dB lower on the average PSNR, and has the best result on one test video (\#15). 

\subsection{Track 3 -- Quality enhancement with x4 SR}

Track 3 is the most challenging track that requires a $\time 4$ super-resolution on the compressed video. Similar to Track 2, we consider the videos that are $\times 4$ upscaled by bicubic algorithm as the unprocessed samples. As shown in Table~\ref{tab:track3}, there are 10 teams that outperforms the state-of-the-art method~\cite{chen2021compressed}. Besides, there are 8 methods that improves the average PSNR by more than 1.0 dB upon the bicubic $\times 4$ videos. The top 3 methods achieve $>$ 1.6 dB PSNR improvement, and forth-rank and the fifth-rank methods also has around 1.5 dB PSNR improvement, in comparison with the bicubic $\times 4$ videos.

In the proposed methods, the top 3 methods (GY-Lab, TaoMC2 and NoahTerminalCV) has comparable performance. The differences on the average PSNR among the top 3 methods are $\leq$ 0.03 dB. The GY-Lab, TaoMC2 and NoahTerminalCV Teams has the best results on 7, 3 and 6 test videos, respectively.  

\subsection{Efficiency, test strategies and training data} 

We report the running time of the proposed methods in Table~\ref{tab:time}. It can be seen from Table~\ref{tab:time} that the GY-Lab Team has the highest time efficiency among the top methods, indicating that it achieves a good trade-off between quality and time efficiency. Besides, as Table~\ref{tab:time} shows, most top methods uses the self-ensemble~\cite{timofte2016seven} to improve the quality performance, and moreover, the GY-Lab Team further uses a temporal ensemble strategy. Some teams also utilize the ensemble of several models, which are trained with different loss functions or at different training epochs. Table~\ref{tab:time} also shows that extra training data of each team. Most top teams trained their models with extra training samples, including REDS~\cite{nah2019ntire}, Vimeo90K~\cite{xue2019video} and the videos collected from YouTube~\cite{youtube}. This indicates that enlarging the training set is beneficial for improving the performance of quality enhancement. Note that, the data in Table~\ref{tab:time} are provided by the participants, so the data may be obtained under different hardware and conditions. Therefore, Table~\ref{tab:time} is only for reference. It is hard to guarantee the fairness in comparing time efficiency.

\section{Teams and methods}

\subsection{TaoMC2 Team}

\begin{figure}[!t]
	\centering 	
 	\includegraphics[width=1.0\linewidth]{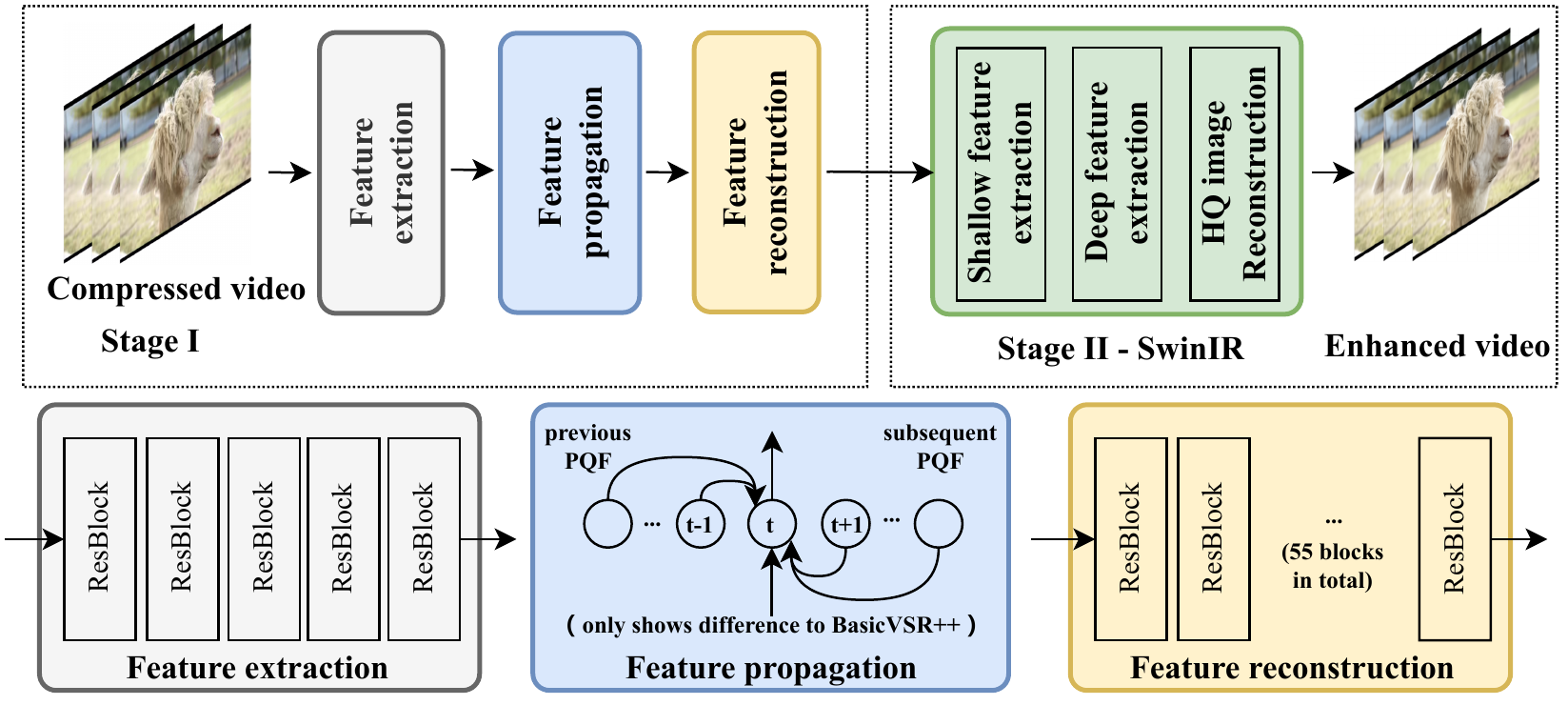}
 	\caption{The two-stage approach proposed by the TaoMC2 Team.}
 	\label{fig:diagram}
 \end{figure}

\textbf{Framework.} The TaoMC2 Team proposes a two-stage approach for quality enhancement on the compressed videos, the pipeline of which is shown in Figure~\ref{fig:diagram}.
In stage I, the network is developed on the top of  BasicVSR++~\cite{chan2021basicvsr++}.
Specifically, they replace the second-order flows in BasicVSR++ by PQF flows~\cite{yang2018multi,guan2019mfqe}.
They also deepen the reconstruction module of BasicVSR++.
Finally, the stage I model is trained over a large-scale high-quality dataset in a progressive manner, which is inspired by the feature progressive sharing and deep supervision training of RBQE~\cite{xing2020early}.
In stage II, they further improve the quality of the enhanced consecutive frames by a state-of-the-art image restoration network, \ie, SwinIR~\cite{liang2021swinir}. This stage helps mitigate severe blurry and further improve the quality upon the single-stage method. Finally, the networks of stage I and II are cascaded for producing the final results.

\textbf{Training.} For stage I of Track 1, they first fine-tune the official pre-trained BasicVSR++~\cite{chan2021basicvsr++} model for 300,000 iterations with the Charbonnier loss~\cite{lai2018fast}, using Adam optimizer with the initial learning rate of $2 \times 10^{-5}$.
They also adopt the Cosine Restart scheduler with the period of 300,000 iterations, and linearly increase the learning rate for the first 10\% iterations.
Besides, they progressively train and converge the model by increasing the number of residual reconstruction blocks from 5 to 55.
Then, the model is fine-tuned with L2 loss for 100,000 iterations.
For stage I of Tracks 2 and 3, they load the pre-trained model of Track 1 and repeat the above training process.
For stage II of Track 1, the image restoration model of SwinIR is fine-tuned over the NTIRE training dataset and the additional 870 videos from YouTube~\cite{youtube}, via the default Charbonnier loss. This SwinIR model is initialized by pre-trained parameters from  \cite{liang2021swinir}, which is trained for RGB image denoising.
Then they jointly fine-tune the overall model with a small learning rate of 10$^{-6}$ using the L2 loss function, over the training datasets. 
For stage II of Track 2, they fine-tune the pre-trained model from Track 1 with Charbonnier loss for 15K iterations. The stage II of Track 3 directly uses the pre-trained model from Track 2 without additional fine-tuning.

\begin{figure}[!t]
	\centering 	
 	\includegraphics[width=1.0\linewidth]{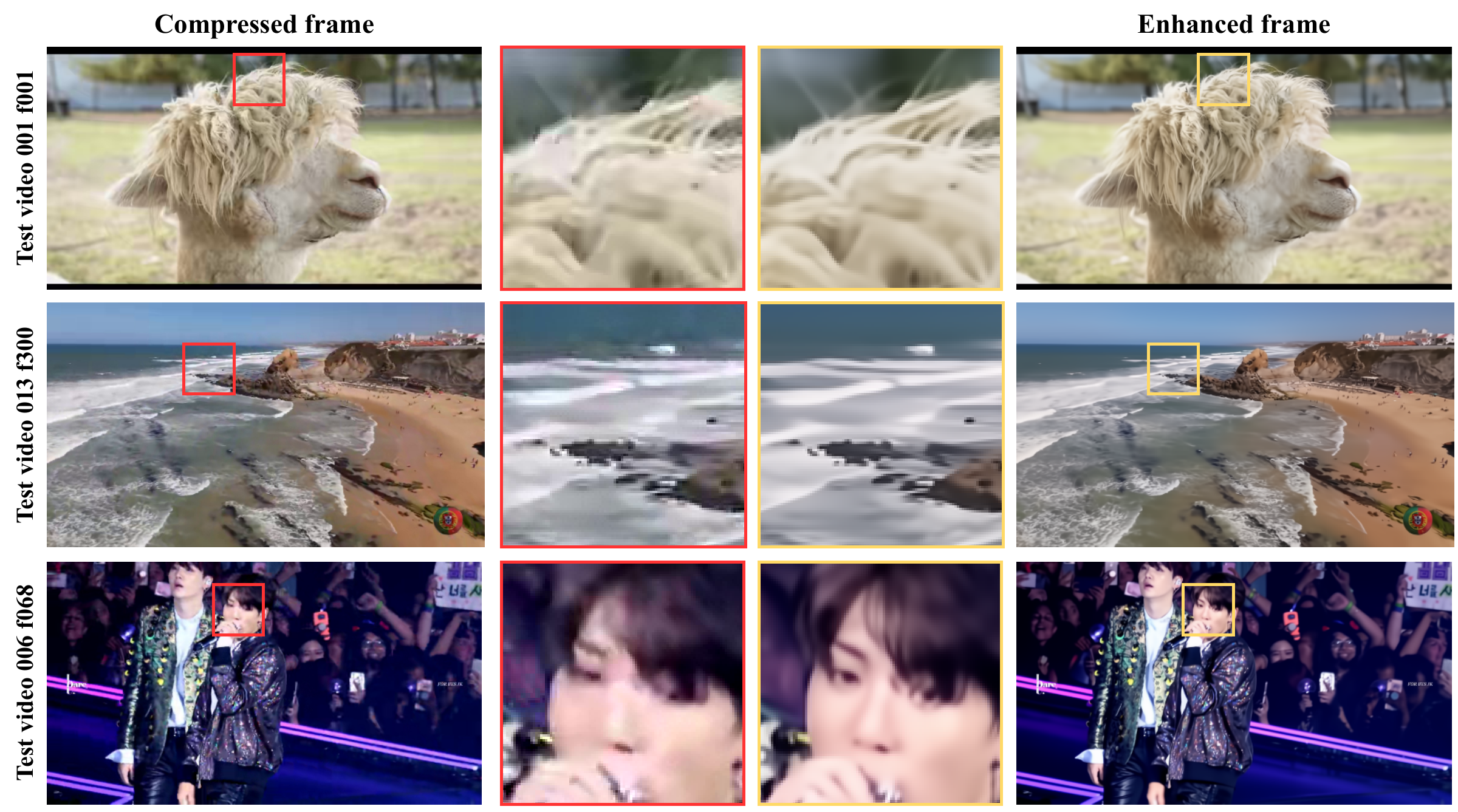}
 	\caption{Visual results of the two-stage approach of TaoMC2.}
 	\label{fig:subjective}
 \end{figure}
 
\textbf{Test.}
For Tracks 1 and 2, all frames of each compressed video are input into the model of stage I to get the enhanced frames.
Then, the enhanced frame is further enhanced by the model of  stage II model.
Moreover, they conduct a 8-set self-ensemble method \cite{timofte2016seven} to
augment the input frames, and then averaging the all enhanced results as the final output.
For Track 3, on the top of self-ensemble, they further conduct a two-set model ensemble for the inference of stage I. Specifically, they average the outputs of two trained models as the final result of stage I, while each model is first conducted above 8-set self-ensemble. Subsequently, the output of stage I is feed into stage II, which is without any ensembles. Figure \ref{fig:subjective} shows the visual results of the TaoMC2 Team. It can be seen from Figure \ref{fig:subjective} that it achieves more details in the blurred regions of video frames, \eg, the hair is much clearer in the enhanced frame. Besides, the output of the proposed method contains less motion blur, compared with the compressed video.

\subsection{GY-Lab Team}

\begin{figure}[t]
\begin{center}
\includegraphics[width=1.0\linewidth]{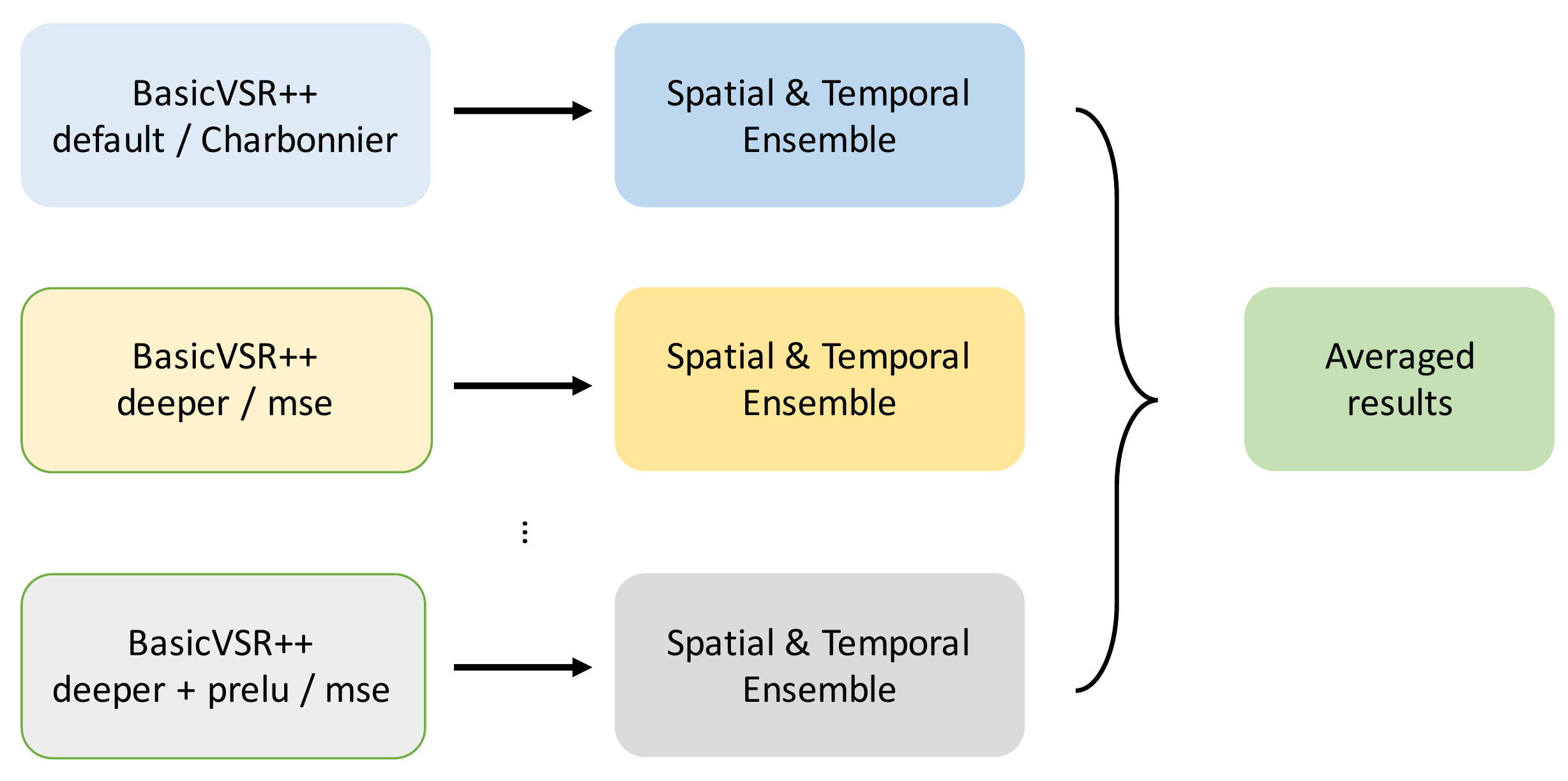}
\end{center}
\caption{The pipeline of the method proposed by the GY-Lab Team in all three tracks. Models under different settings (structure, loss functions) are first trained independently. They apply both spatial and temporal ensemble on each model and average the outputs of all models as the final result. }
\label{fig:overall}
\end{figure}

\textbf{Framework.} The method of the GY-Lab Team is built on BasicVSR++\cite{chan2021basicvsr++}. Figure~\ref{fig:overall} shows the pipeline of the proposed method. Inspired by \cite{lin2022revisiting}, an enlarged model combined with proper training strategies is expected to have noticeable improvements over the baseline. On the one hand, they perform two modifications on BasicVSR++ to improve its capacity. To be specific, since the modules of feature extracting and upsampling are far weaker than that of feature propagation, they increase the number of residual blocks in these two modules. Besides, enhanced activation functions (\eg, SiLU\cite{elfwing2018sigmoid}) are proven to be effective according to \cite{lin2022revisiting}. For the sake of training efficacy, they replace Leaky ReLU in BasicVSR++ with PReLU\cite{he2015delving} and verify its effectiveness. On the other hand, they adopt different loss functions (\eg, MSE loss and Charbonnier loss~\cite{lai2018fast}) to supervise model training so as to obtain results with different aspects of advantages, which is beneficial to subsequent ensemble.

\textbf{Training. } For improving the generalization ability of the proposed method, they employ the videos from REDS\cite{nah2019ntire}, Vimeo90K\cite{xue2019video} and YouTube as training samples, in addition to the NTIRE training set. In Track 1, they adopt open-sourced BasicVSR++\cite{chan2021basicvsr++} as the baseline as well as the pre-trained model. In terms of data augmentation, they follow the settings in BasicVSR++ except the patch size is increased from $256$ to $384$. Besides, the model is enlarged by two modifications mentioned above to increase the capacity. Low-precision training (fp16) is adopted to enable the training of larger model with bigger patch size. Different loss functions (e.g., MSE loss and Charbonnier loss~\cite{lai2018fast}) are leveraged to train models independently for results of diverse characteristics. Each model is first trained for $300,000$ iterations with Adam\cite{kingma2014adam} optimizer (initial learning rate of $4\cdot 10^{-5}$ and batch size of $8$). After that, another training with $100,000$ iterations is performed with an initial learning rate of $4\cdot 10^{-6}$.
In Track 2 and Track 3, they utilize the models trained in Track 1 as the pre-trained models. Similar training strategy is adopted for these two tracks. Note that for the models of different super-resolution scales, most layers in the pre-trained models have identical structures except the modules of feature extracting and upsampling. Therefore, they set a larger initial learning rate of $10^{-4}$ for these layers to encourage faster convergence.

\begin{figure}[t]
\begin{center}
\includegraphics[width=1.0\linewidth]{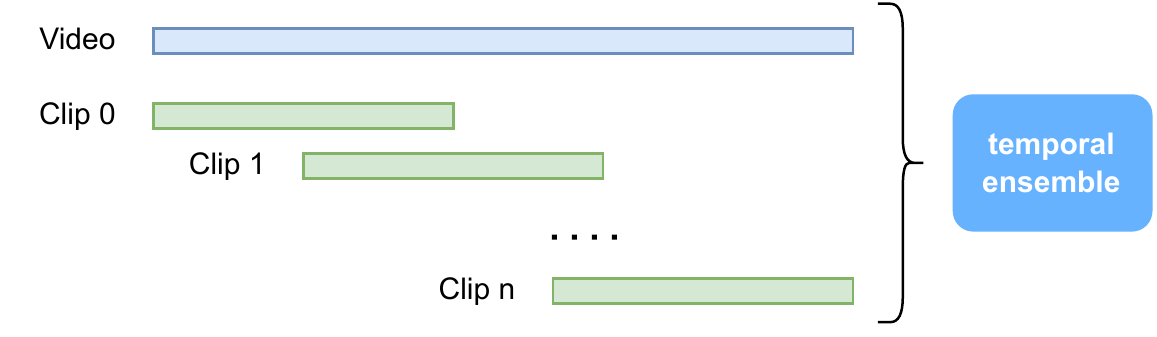}
\end{center}
\caption{Details of the temporal ensemble strategy of the GY-Lab Team. Each video is split into several clips with overlap and the average of these outputs is taken as the ensemble result. }
\label{fig:ensemble}
\end{figure}

\textbf{Test. } In the first stage, they utilize spatial and temporal self-ensemble strategies to improve the final results. For spatial ensemble, they feed augmented input frames independently to the network, including horizontal flip, vertical flip and rotation, and use the average outputs as prediction. This brings $0.18\sim0.20$ dB improvement on PSNR. As for temporal ensemble, they split each video into overlapped clips and obtain average results as shown in Figure \ref{fig:ensemble}. Since short clip length causes unexpected performance drop, temporal clip length is restricted to $200$, which results in $0.01\sim0.02$ dB PSNR improvement.
Besides self-ensemble, they also use conventional multi-model ensemble strategy to further improve the result. Among the three tracks, they choose several models trained with different hyper-parameters and fuse the results by averaging. However, the multi-model ensemble is time-consuming and cannot always get a stable improvement. It performs better on Track 3 than Tracks 1 and 2.

\subsection{HIT\&ACE Team}

Model fusion is a commonly used strategy for boosting performance, but averaging the results of exactly the same models which are re-trained for multiple times only provides marginal improvement. In order to obtain models that are complementary to each other, they propose training models with different architectures and sizes, and then fuse the results by weighted average.
Specifically, they use BasicVSR++~\cite{chan2021basicvsr++} as the backbone architecture.
For track1, they train two models. In one of them, they replace the five reconstruction residual blocks with four transformer blocks~\cite{zamir2021restormer}. In the other model, they increase the number of channels to 256 and enlarge the number of the residual blocks in each recurrent step to 30 (denoted as large-BasicVSR++). The results of the two models are averaged with weights of 0.3 and 0.7, respectively.
In track2, they only use a large-BasicVSR++ model. In track3, they train one BasicVSR++ model and one large-BasicVSR++ model. Then, they average the results with weights of 0.7 and 0.3, respectively.

\subsection{NoahTerminalCV Team}

\begin{figure}[t]
    \centering
    \includegraphics[width=\linewidth]{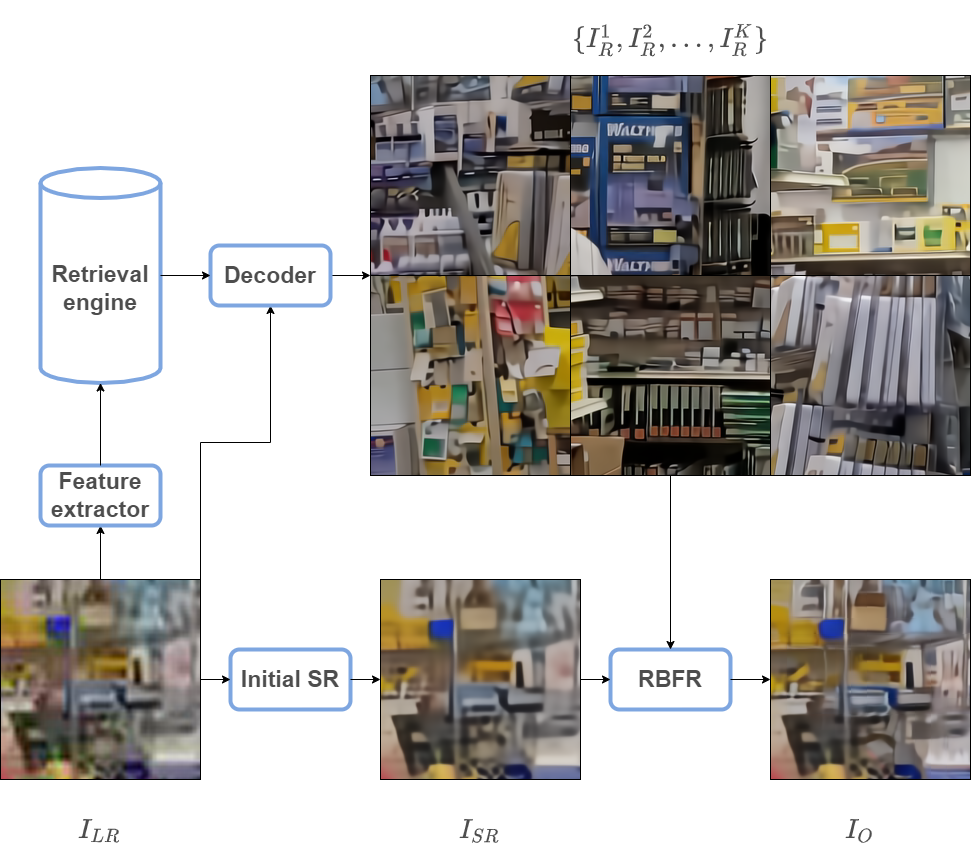}
      \caption{The proposed method of the NoahTerminalCV Team.}
  \label{fig:rbfr}
\end{figure}

The proposed method of the NoahTerminalCV Team consists of two subsequent stages. Firstly, they perform an initial super-resolution using an ensemble of feed-forward multi-frame neural networks.
The second step is called reference-based frame refinement. They find the top K similar images for each low-resolution input frame from the external database. Then, they run a matching correction step for every patch on this input frame to perform a global alignment of reference patches. As a result, we have the $I_{SR}$, which comes from the first stage, and a set of globally aligned references $\{I_{R}^1, I_{R}^2,..., I_{R}^K\}$. Finally, they are processed with the RBFR network ($N_{RBFR}$) to handle residual misalignments and to properly transfer texture and details from the reference images to initially super-resolved output $I_{SR}$, and then the final output $I_{O}$ can be obtained. The proposed framework is illustrated in Figure~\ref{fig:rbfr}.

\subsubsection{Initial super-resolution}
Many different networks have been tried to perform the initial super-resolution. The key ideas which are followed to design a network are briefly introduced as follows:

\begin{enumerate}
    \item As a general pipeline, they follow the NoahBurstSRNet~\cite{bhat2021ntire} which is a window-based neural network. It takes a low-resolution image to enhance (key-frame) and also N additional frames from the video. The design of this network architecture allowed us to use N=16 during training and N=128 during inference;
    \item The original convolutional blocks in NoahBurstSRNet are replaced with SwinIR~\cite{liang2021swinir} modules;
    \item The original alignment module in NoahBurstSRNet is based on PCD~\cite{wang2019edvr} which is a pyramid alignment with deformable convolutions ~\cite{dai2017deformable}. To improve the alignment procedure, they employ the ideas from ~\cite{chan2021basicvsr++} and ~\cite{wang2019edvr} and build a new alignment module which is called Pyramid Flow-Guided Deformable Alignment (PFGDA). The main difference from PCD is the usage of optical flow estimated by a pretrained GMA~\cite{jiang2021learning} network as a residual for an offset prediction block.
    \item They employ a bitstream parser~\cite{rao2020bitstream} to extract codec information from the input videos. For each of the frames, it extracts various statistics, including average motions, QP values, block sizes, etc. In total, for each input frame, they build a vector of 127 numbers. Then, to supply the network with this information, they replace LayerNorm~\cite{ba2016layer} layers in Swin Transformer blocks with Adaptive Layer Normalization which is similar to AdaIN~\cite{huang2017arbitrary}.
    \item To increase the capacity of the network without affecting training and inference speed, they added a trainable Product Key Memory (PKM) layer~\cite{lample2019large} before the last Swin Transformer block.
\end{enumerate}

The super-resolution network is trained using a pixel-wise L1 objective on the full input images without cropping. The training of one network until convergence takes about 7 days using 64 NVIDIA Tesla V100 GPUs. Using a bigger batch size typically leads to better performance on the validation dataset.

\subsubsection{Reference-based frame refinement (RBFR)}
To employ the reference-based refinement strategy, they first build a retrieval engine. The database consists of 1,400,000 images of size $960\times512$ sampled from a training dataset. The naive way of storing the database would take 2 TB which is challenging for practical usage. Therefore, they train an autoencoder to compress the database.

\textbf{AutoEncoder.}
The typical autoencoder consists of two parts: Encoder $N_E$ and Decoder $N_D$. The image is firstly processed by the encoder to obtain a latent representation $z = N_E({I_{HR}})$ and then the decoder can be used to reconstruct the original input $\widehat{I}_{HR} = N_D(z)$. To improve the reconstruction quality of the AutoEncoder, they supply the decoder with additional $I_{LR}$ input. Therefore, in this case, the reconstruction is obtained as follows: $\widehat{I}_{HR} = N_D(z, I_{LR})$. This approach can improve the reconstruction quality by 0.5 dB.
The autoencoder is trained using the L1 objective. Also, they use a pair of unaligned images $I_{LR}, I_{HR}$ from the same video for training to improve the robustness.

\textbf{Feature Extractor.}
To build a retrieval engine, they train a feature extractor network that takes a low-resolution image $I_{LR}$ and represents it as a feature vector. They use a contrastive learning~\cite{chen2020simple} framework to train the feature extractor. For positive samples, they use two random frames from the same video, while for the negative samples we employ frames from different videos. The Resnet-34~\cite{he2016deep} architecture is utilized as the feature extractor.

\textbf{Retrieval Engine.}
After compressing the database of images using the trained Encoder, obtaining latent representations, and representing all low-resolution versions as a feature vector of size 1000 extracted from the trained Feature Extractor, they build an index using the HNSW~\cite{DBLP:journals/corr/MalkovY16} algorithm from the nmslib~\cite{DBLP:conf/sisap/BoytsovN13} library. This algorithm allows searching for the top K nearest neighbors in the database.

\textbf{RBFR.}
Finally, they train a network $N_{RBFR}$ that takes the result of initial super-resolution $I_{SR}$ and top K similar images from the database $\{I_{R}^1,I_{R}^2,...,I_{R}^K\}$. The network produces the final prediction $I_O$. We train $N_{RBFR}$ through the L1 objective between $I_O$ and $I_{HR}$. As a $N_{RBFR}$, the NoahBurstSRNet~\cite{bhat2021ntire} architecture is used, since it effectively handles small misalignments and can properly transfer information from reference non-aligned images.

\subsubsection{Test}

During the inference, in order to upscale the key-frame $I_{LR}^i$ we put it to the initial super-resolution network together with additional frames $I_{LR}^{i-1}, I_{LR}^{i+1}, I_{LR}^{i-2}, I_{LR}^{i+2}...$. The number of additional frames during the inference is set to 128.
For RBFR, they first obtain top K (typically 16) similar images using the retrieval engine. Then, the inference is done in a patch-wise manner. They extract a patch from the $I_{SR}$ and use the Template Matching~\cite{briechle2001template} to perform a global alignment and find the most similar patches on the images $\{I_{R}^1, I_{R}^2,..., I_{R}^K\}$. Then, they are fed to the $N_{RBFR}$ to generate the final result.

\subsection{BOE-IOT-AIBD Team}

\begin{figure}[t]
  \centering
  \includegraphics[width=\linewidth]{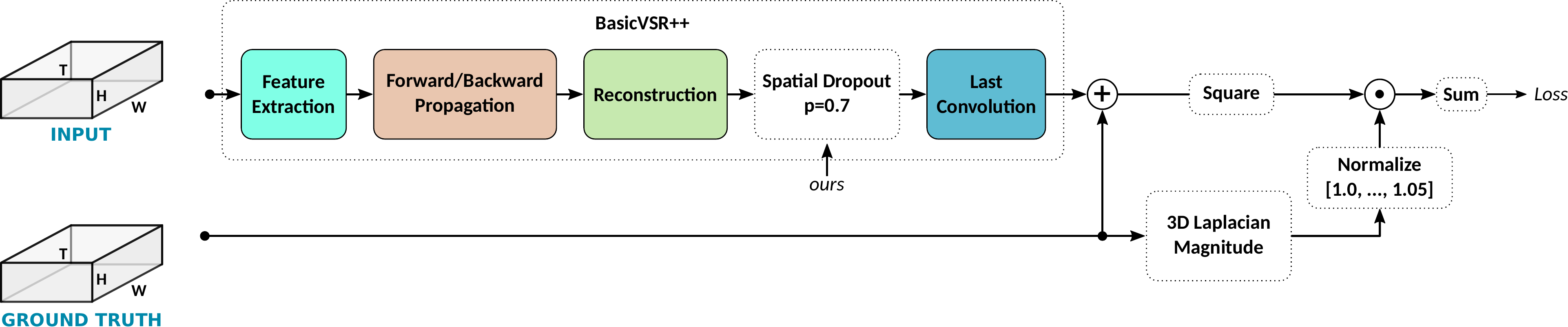}
  \caption{Proposed framework of the BOE-IOT-AIBD Team. \label{fig:boe}}
\end{figure}

Given the success of the BasicVSR++ model in NTIRE 2021~\cite{chan2021basicvsr++}, they focus on the training strategies using this model. Based on the recommendations in \cite{RevRCAN, ConvNEXT, kong2021reflash}, they applied several strategies including: adding spatial dropout~\cite{tompson2015efficient} before the last layer of the model, using large batch size ($240$) and increasing the batch volume focused mostly on an increased number of frames (from $75$ to $120$). The major difficulty is the memory requirements to run training on large batches and for this purpose they use the automatic mixed--precision library APEX\footnote{\url{https://nvidia.github.io/apex/}} from NVIDIA. Finally, they use a loss function that weighted the MSE error per pixels based on the 3D Laplacian magnitude, increasing the relevance of pixels along 3D edges (spatial and temporal). Figure~\ref{fig:boe} illustrates the proposed framework of the BOE-IOT-AIBD Team.

\subsection{ZX\_VIP Team}

The ZX\_VIP Team trains the BasicVSR++ network~\cite{chan2021basicvsr++} by the following steps.
In pre-train phase, they train the model on the REDS dataset~\cite{nah2019ntire} with Adam optimizer and the CosineRestart scheme. The initial learning rate of the main network and the flow network are set to $10^{-4}$ and $2.5\cdot 10^{-5}$, respectively. The total number of iterations is 200,000, and the weights of the flow network are fixed during the first 5000 iterations. The batch size is 16 and the patch size of input low resolution frame is $64\times 64$. They use the Charbonnier loss function, and use the pre-trained SPyNet~\cite{ranjan2017optical} as the flow network. The number of residual blocks for each branch is set to 25 and the number of feature channels is 128. Then, in the training phase, they train the model on the NTIRE training set. They adopt the Adam optimizer and the step scheme. The initial learning rate of the main network and the flow network are set to $2\cdot 10^{-5}$ and $5\cdot 10^{-6}$, respectively. The total number of iterations is 600,000, and the weights of the flow network are fixed during the first 5000 iterations. The batch size is 12 and the patch size of input LR frame is $48\times 48$. They also use the Charbonnier loss and load the above pre-trained network. They also perform data augmentation during the training phase, \ie, horizontal flip and vertical flip.

\subsection{OCL-VCE Team}

\begin{figure}[t]
    \centering
    \includegraphics[width=1\linewidth]{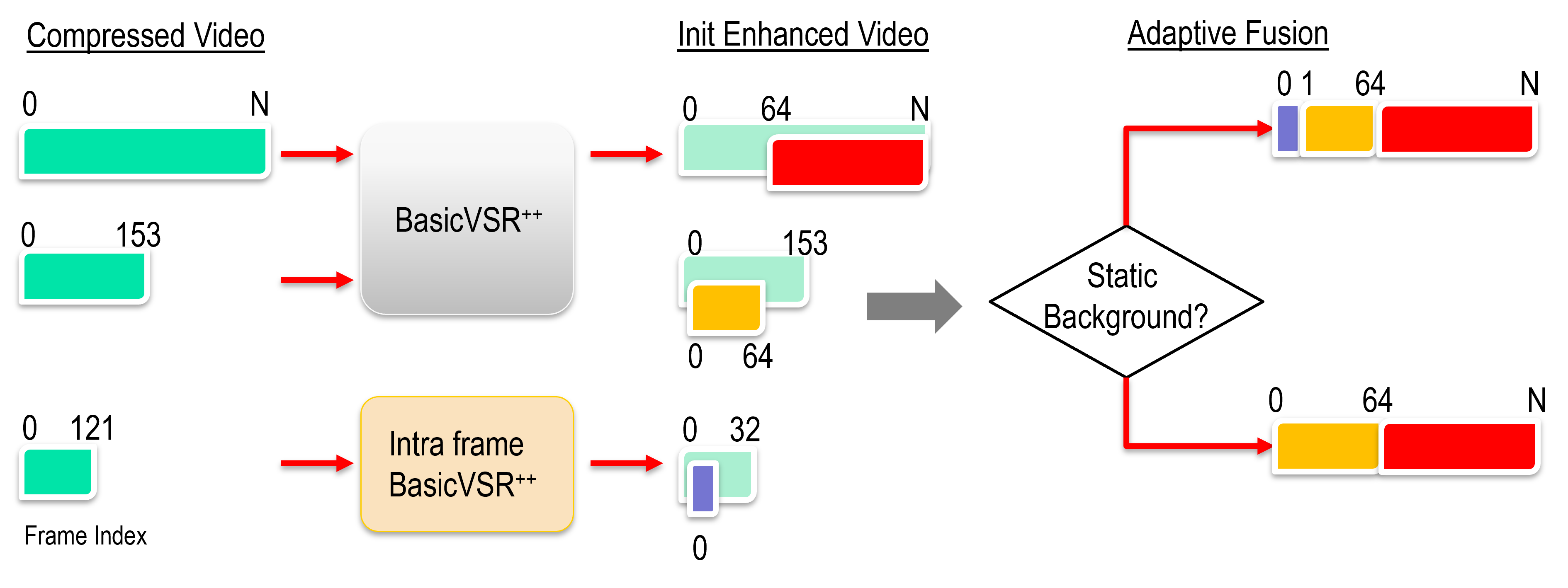}
    \caption{The framework proposed by the OCL-VCE Team. The static video is detected via the gradient of the averaged frames.}
    \label{fig:general}
\end{figure}

The OCL-VCE Team propose training the model for I-frames on the video clips with 30 frames, that are cut from the original training videos and encoded by the LDP configurations. Then, they fine-tune the intra-frame BasicVSR++ on the video clips with the first frame encoded as I-frame. They observed  improvement on I-frames except for few sequences at high frame-rate and with slow motion (\ie, video 226, 227). These cases can be separated by comparing the gradient of the averaged frame with a given threshold. The averaged frame is calculated as
\begin{equation}
    \bar{f} = \sum_{i}^{i*m < N} f_{i*m},
\end{equation}
where $m$ is a scaling factor, which is set as $m=4$ for the videos with the frame-rates less than 30 fps and as $m=8$ for those with the frame-rates greater than 30 fps. Then, its gradient is compared with a given threshold
\begin{equation}
    \nabla{f} = ||\nabla_x f|| + ||\nabla_y f|| < \tau,
\end{equation}
where $\nabla_x, \nabla_y$ denote the gradients in the horizontal and vertical directions, respectively. $\tau$ is a given threshold of 22,500. $\tau$ can be normalized based on the number of pixels. Therefore, they propose fusing multiple enhanced results with difference input frames based on the video context, as shown in Figure~\ref{fig:general}.

\subsection{Trick collector Team}

\begin{figure}[t]
    \begin{center}
       \includegraphics[width=1\linewidth]{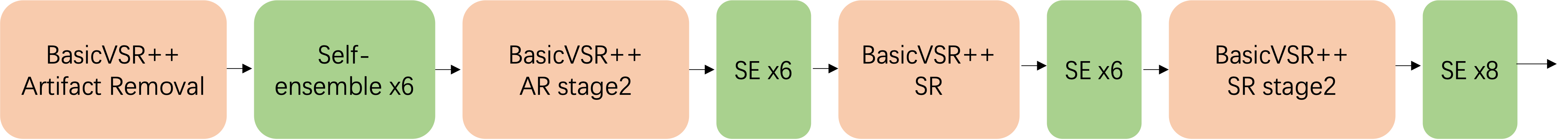}
    \end{center}
       \caption{The method for Track 2 of the Trick collector Team.}
    \label{fig:track2}
\end{figure}

\begin{figure}[t]
    \begin{center}
       \includegraphics[width=1\linewidth]{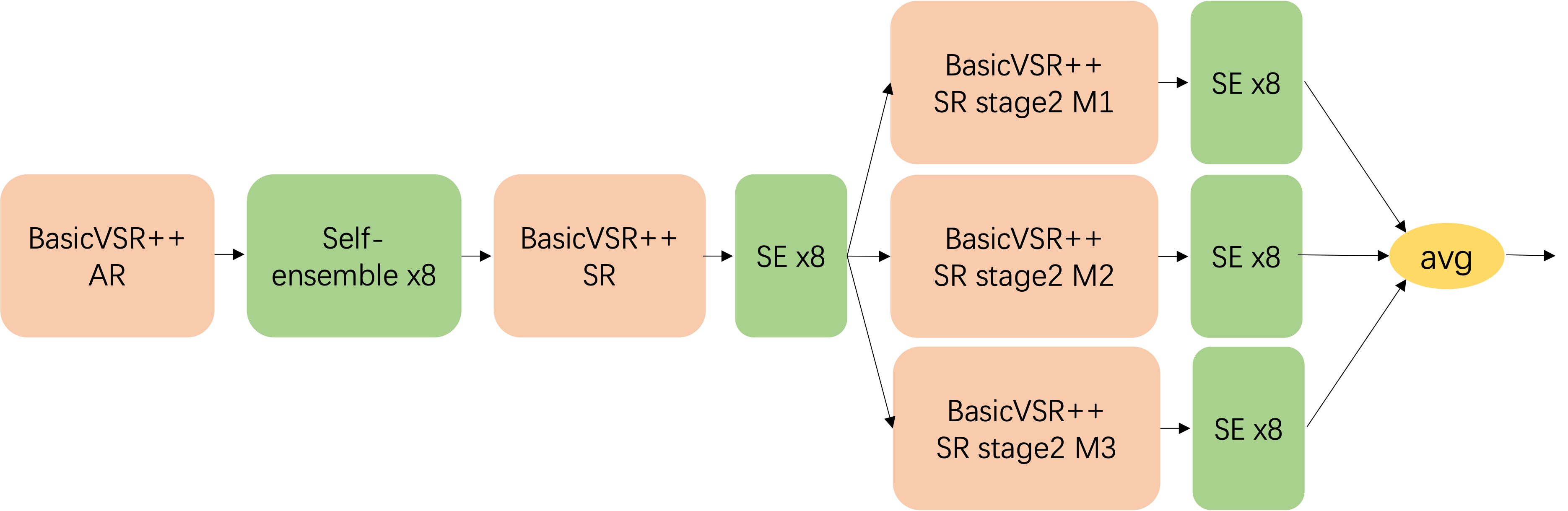}
    \end{center}
       \caption{The method for Track 3 of the  Trick collector Team.}
    \label{fig:track3}
\end{figure}

The Trick collector Team proposes a separated framework for video quality enhancement. Their contributions can be summarized as follows. First, they divide video quality enhancement into two sub-tasks: artifact reduction and super resolution. For each sub-task, they use an one-stage or two-stage model. Specifically, for Track 2, they use the two-stage models on both artifact reduction and super resolution. For Track 3, they use an one-stage model on artifact reduction and a two-stage model on super resolution. The baseline model of each stage is BasicVSR++~\cite{chan2021basicvsr++}. Besides, they find  that averaging results of multiple models on the last stage of Track 3 is helpful. Therefore, they train three models with different hyper-parameters on the last stage of Track 3 and average their results as the final results.
Besides, they also consider self-ensemble in the training phase. Typically, self-ensemble~\cite{timofte2016seven} is only used in the test phase. However, in the separated framework, self-ensemble can be used in each stage of each sub-task. In order to make the model better adapt to the results of self-ensemble, self-ensemble should also be considered in the training phase. Specifically, in the training phase, they use $256\times 256$ RGB patchs from the training set as input, and augment them with random horizontal flips and 90 rotations. All of our models are optimized by the Adam optimizer with mini-batches of size 1, with the learning rate initialized to $10^{-4}$ using the cosine annealing restarts~\cite{loshchilov2016sgdr} strategy. They use MSE loss (L2 loss) as the loss function.

\subsection{XPixel Team}

\begin{figure}[t]
\begin{center}
\includegraphics[width=1.0\linewidth]{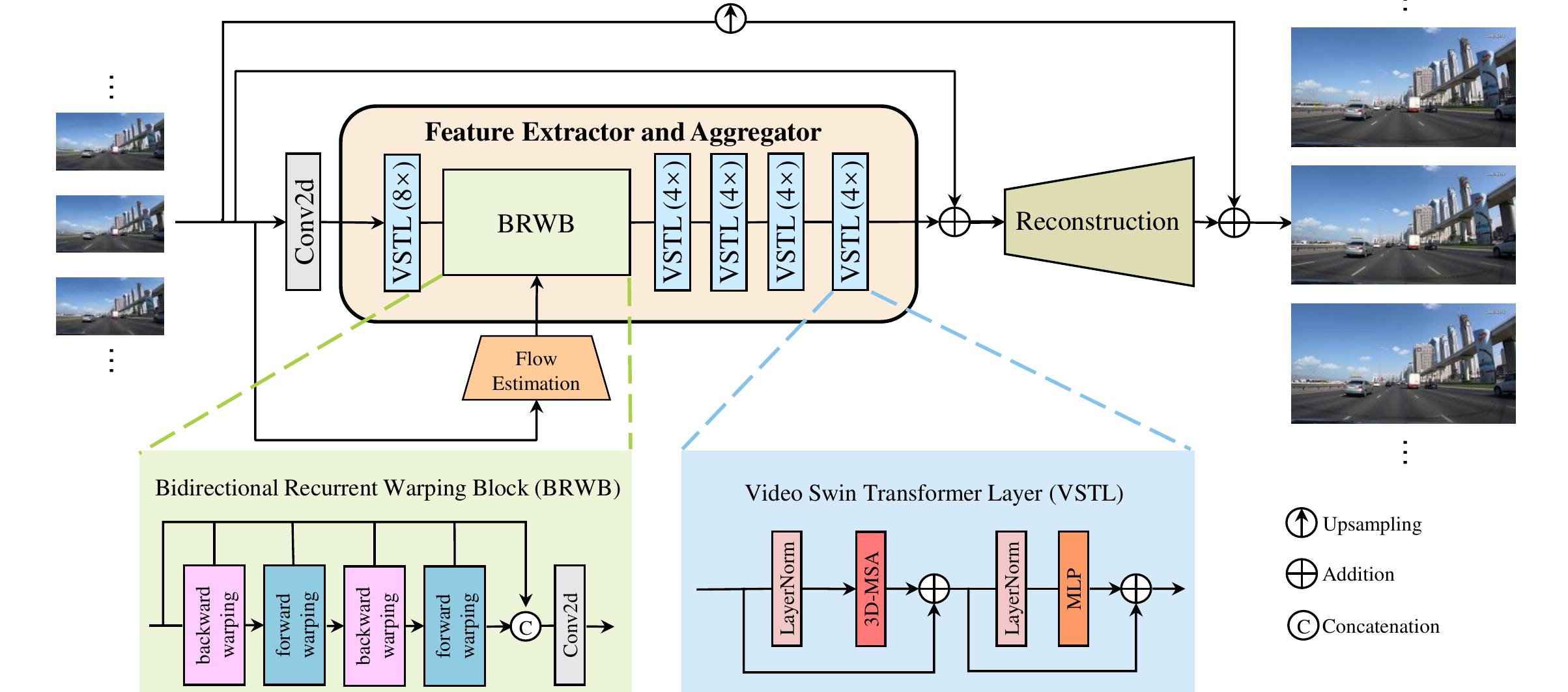}
\end{center}
\caption{Overview of the proposed BRT framework of the XPixel Team. This figure shows an example with three input frames.}
\label{fig:EVRT}
\end{figure}

Inspired by Video Swin Transformer~\cite{liu2021video} and BasicVSR++~\cite{chan2021basicvsr++}, the XPixel Team designs a Bidirectional Recurrent Transformer (BRT) by combining the Video Swin Transformer block and BasicVSR++. These modules effectively increase the implicit and explicit interaction of information between multiple frames and perform well on the challenge validation dataset. The overall framework is shown in Fig.~\ref{fig:EVRT}. The model recovers $N$ high-resolution frames based on $N$ low-resolution frames. First, to enhance the interaction of information between multiple frames, several Swin Transformer layers are used to align frames in an implicit way. Then, to handle large motions between frames, BasicVSR++ is used to align, aggregate and propagate features explicitly. Some other Swin Transformer layers are then added to further refine features. Finally, a reconstruction module is employed to produce high-resolution frames based on the processed features.
The key components of the proposed method is highlighted in the following.

\textbf{Video Swin Transformer Block.}
They use several Video Swin Transformer blocks to align frames with 3D multi-head self-attention (3D-MSA) implicitly in the feature level. 3D-MSA focuses on extracting spatio-temporal global information from frames. The overview of Video Swin Transformer blocks is shown in Figure~\ref{fig:EVRT}.

\textbf{Bidirectional Recurrent Warping Block.}
Considering the effectiveness of BasicVSR++~\cite{chan2021basicvsr++} in terms of alignment and aggregation of inter-frame information, they use it to align and propagate the features. In this module, features of frames are propagated and warped 4 times in the grid structure explicitly. After that, features of each pipeline are aggregated to produce the high-resolution frames.

\textbf{Extended Dataset.}
Overfitting may affect the generalization ability of the model. When training our network with only official training dataset, they encounter a severe overfitting issue. To overcome it, they collect more than 2000 HD video clips from YouTube. The paired dataset is synthesized according to the degradation script provided by organizers.

\textbf{Training and Testing Techniques.}
In the training process, inputting more training frames improves the performance. Therefore, they design a model that can be trained with 100 frames by checkpoint optimization. In the testing process, they use patch-based testing method in order to input more frames. This strategy further improves the performance of the proposed method.

\subsection{OREO Team}

\begin{figure}[t]
	\centering
	\includegraphics[width=\linewidth]{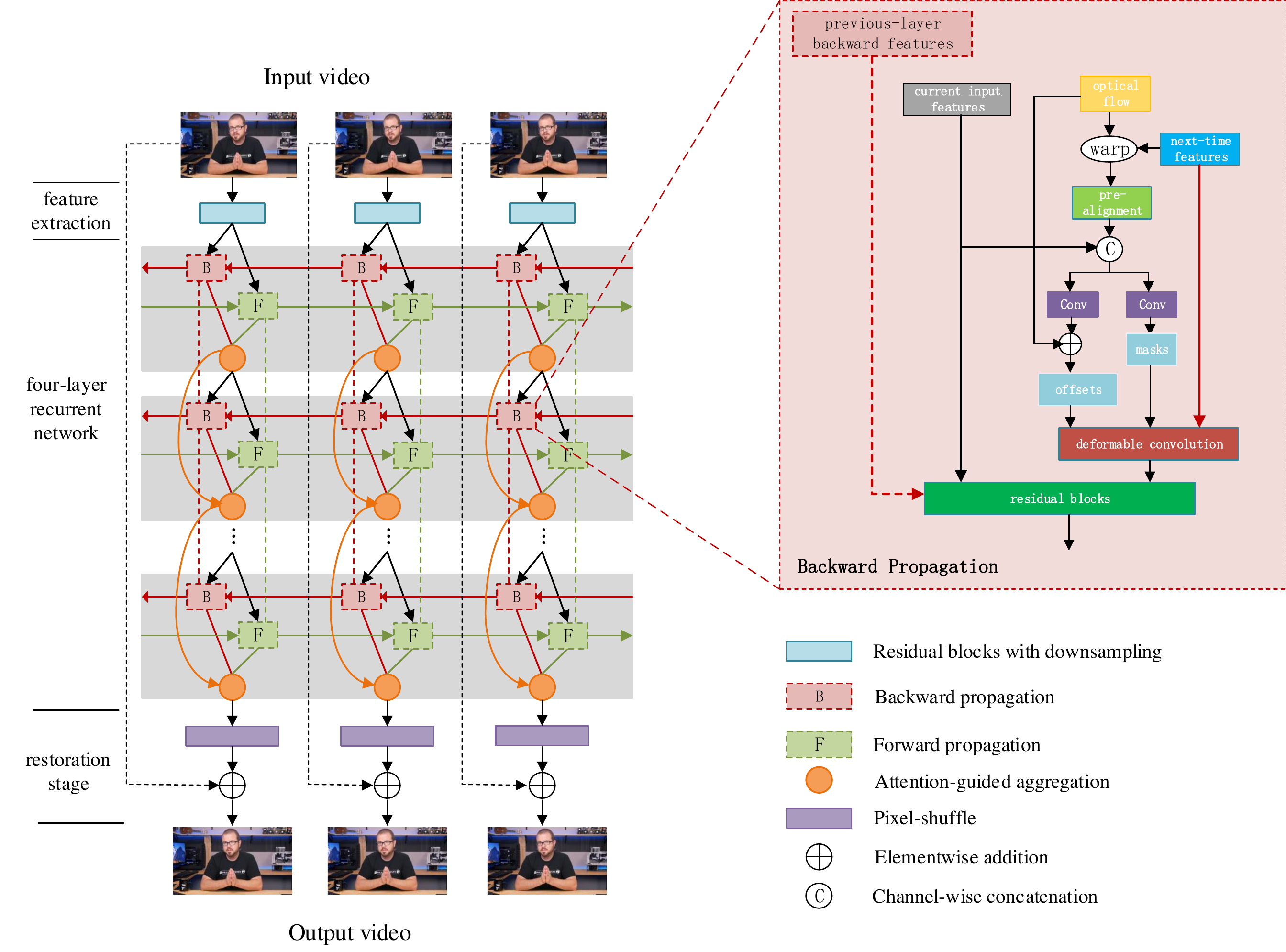}
	\caption{An overview of the architecture proposed by the OREO Team. For simplicity, the connections from shallow feature extraction to the attention-based aggregation block are omitted in the figure.} \label{oreo_fig1}	
\end{figure}

For Track 1, the OREA Team proposes a bi-directional recurrent network with attention-guided aggregation for compressed video quality enhancement, based on the architecture of BasicVSR++~\cite{chan2021basicvsr++}. As shown in the Figure~\ref{oreo_fig1}, the pipeline of the proposed method mainly consists of three parts: feature extraction, a recurrent network with bi-directional propagation equipping with attention blocks and a restoration stage.
		
First, a series of residual blocks are utilized for shallow feature extraction, which allows all subsequent operations be performed on the feature level, such as feature alignment and propagation. Then, a four-layer recurrent network are applied to propagate and supplement inforamtion between neighboring frames. Here, each layer contains a forward branch and a backward branch, in which two branchs first carry out feature aligment and propagation independently, and then their features are selective aggregated to refine information and for the next layer. After this, information from all stages of the recurrent framework are aggregated to restore the final enhanced image for each video frame.
		
\textbf{Bi-directional Recurrent Network}. In order to enjoy the complementary information from neighboring frames, they design a parallel multi-layer bidirectional network which is denoted in Figure~\ref{oreo_fig1}. For each layer, the intermediate features are propagated backward or forward independently. Take the backward propagation as an example. Specifically, the next-time frame features are first aligned to the current features, and then the aligned results are further combined with the current features into residual blocks to take advantage of information from neighboring frames. The outputs are delivered to the previous frame and performed alignment and combination sequently in a backward manner. Here, they employ the flow-guided deformable alignment mechanism as in BasicVSR++. Different from the BasicVSR++, they apply an attention-guided fused block and residual blocks to discriminatively aggregated the backward and forward information to the current features, outputs of which are more informative and further set as the input of the next layer of network. Consequently, the aligned block aims to make the neighboring frame aligned with the enhanced features from the previous layer of the current frame, instead of the initial shallow features. The bidirectional propagation allows the information flow from the first and last frames to the current frame, enhancing the feature expressiveness.
		
\begin{figure}[t]
	\centering
	\includegraphics[width=\linewidth]{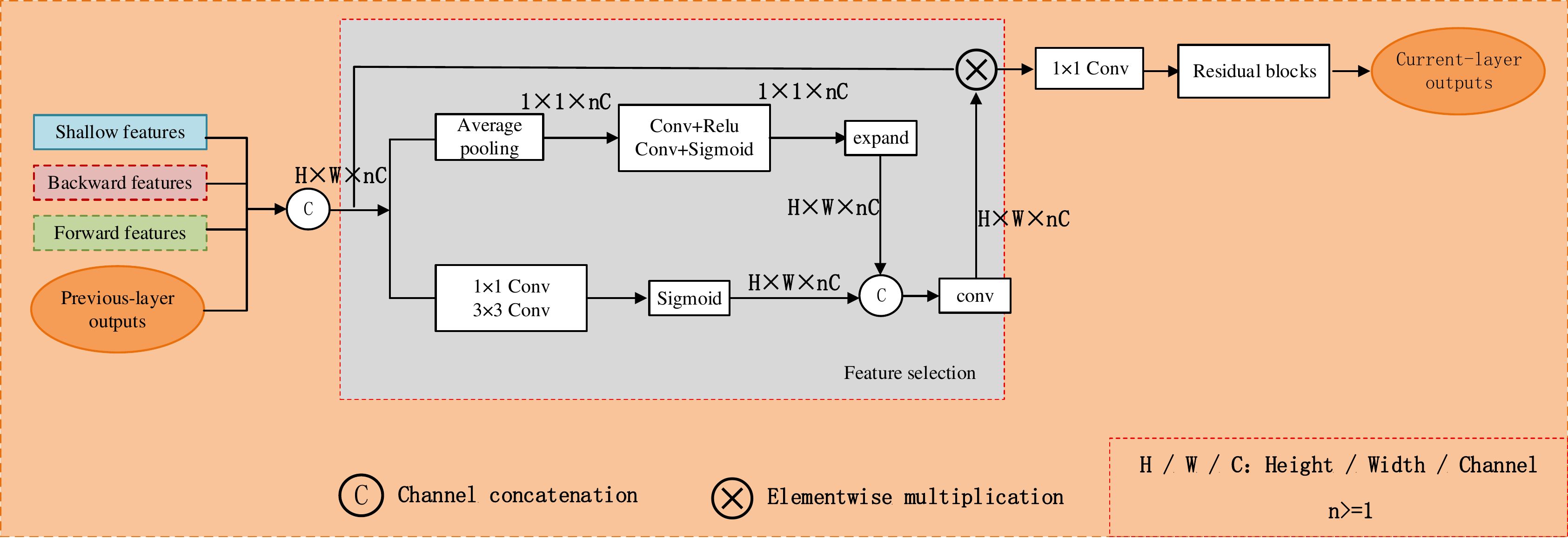}
	\caption{Attention-based aggregation block in the method of the OREO Team.}	
	\label{oreo_fig2}
\end{figure}

\textbf{Attention-Guided Aggregation}. Take the features from backward and forward propagation of the current layer, and those from all previous layers of the current frame as input, they apply an attention block to discriminatively emphasize the informative components and restrain the useless ones, boosting the feature characterization ability. Generally, the accuracy of optical flow and offset prediction greatly affects the alignment, which will further impose a negative impact on the current frame features. They take a combination of channel and spatial attention mechanism to adaptively select the effective and appropriate information for subsequent feature fusion between frames, which helps mitigate the impact of the inaccurate predictions of the optical flow and offset. On one hand, they take an average pooling operation and two convolutions with a Sigmoid for input as in RCAN, to generate a group of weights for each feature channel of input. One the other hand, they apply a 1$\times$1 and a 3$\times$3 convolution to input to produce weights for each spatial point of input. Then, two-branch weights are combined as a new weight set, which further are multiplied by the input to enable feature selection and refinement. The mechanism of screening before fusion is conducive to the effective aggregation of features from backward and forward direction with the current features. The attention-based aggregation block is shown in Figure~\ref{oreo_fig2}.

\subsection{HMSR Team}

\begin{figure}[t]
\centering
\includegraphics[width=\linewidth]{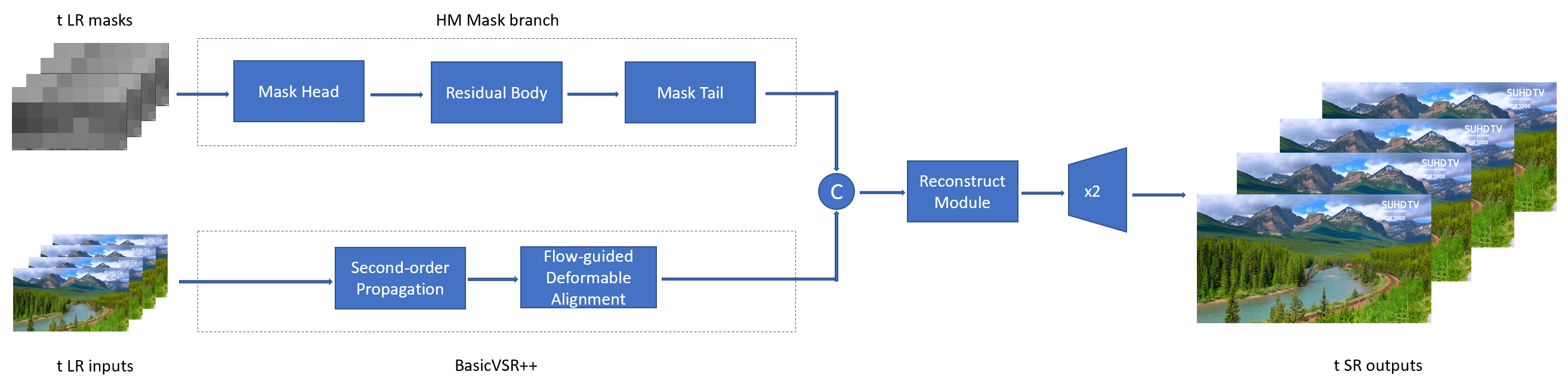}
\caption{The proposed method of the HMSR Team.}\label{hmsr-architecture}
\end{figure}

Based on BasicVSR++\cite{chan2021basicvsr++}, the HMSR Team proposes an HM-mask branch for the reconstruction module. They apply the HEVC test model (HM) to extract the information of block division of low resolution videos, and then introduce them into the network as a prior to provide additional information for compressed video super-resolution, thereby improving the final performance. The proposed method is illustrated in Figure~\ref{hmsr-architecture}.

\subsection{UESTC+XJU CV Team}

The UESTC+XJU CV Team uses BasicVSR++~\cite{chan2021basicvsr++} in Track 1. The number of residual blocks for the initial feature extraction is set to 5. The number of residual blocks for each branch is set to 25. The number of feature channels is 128. In the training process, the raw and compressed sequences are cropped into $256\times 256$ patches as the training pairs, and the batch size is set to 4. Meanwhile, the 15 compressed frames are used as inputs.
They also adopt flip and rotation as data augmentation strategies to further expand the dataset.
The model is trained by Adam optimizer~\cite{kingma2014adam} with $\beta_{1}=0.9$, $\beta_{2}=0.999$ and $\varepsilon=10^{-8}$ for $6\cdot 10^{5}$ iterations.
The learning rate is initially set to $10^{-4}$ and retained throughout training.
Following~\cite{chan2021basicvsr++}, they use Charbonnier loss~\cite{charbonnier1994two} as the loss function and use  pre-trained SPyNet~\cite{ranjan2017optical} as the optical flow network. They take the 165 video frames as inputs to
explore long-range temporal information for restoration. 

\subsection{TBE Team}

\begin{figure}[t]
\begin{center}
\includegraphics [width=1\linewidth]{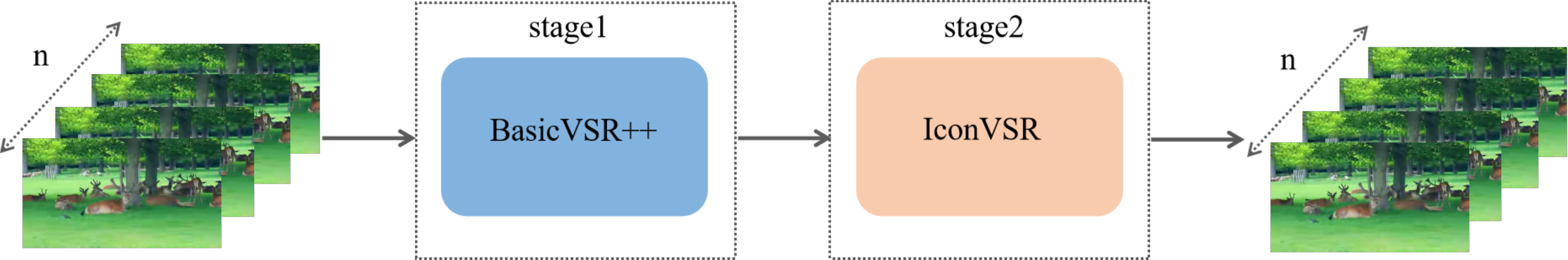}
\caption{The proposed framework of the TBE Team.}\label{tbe}
\end{center}
\end{figure}

The proposed architecture of the TBE Team is composed of two cascaded stages. As shown in Figure~\ref{tbe}, they apply a recurrent video super-resolution framework BasicVSR++~\cite{chan2021basicvsr++} in stage 1. Then, the stage 2 includes a video quality enhancement model IconVSR~\cite{chan2021basicvsr} without the upsampling layers. Two stages are trained individually in two phases. In the first phase, they feed the low resolution frames into the model of stage 1 (BasicVSR++) and obtain the best super-resolved outputs. Then, they take the video super-resolution results as the input to stage 2 for training the IconVSR model to acquire the final results.

During the training phase, the total data we used are 331 video sequences. They choose 307 videos for training and 24 videos for evaluation. For the training of BasicVSR++, they randomly extracted 30  consecutive frames from each low resolution sequence. The $128\times 128$ low resolution patches are randomly cropped from each of the 30 frames as the low resolution inputs. At the same time, the corresponding $256\times 256$ high resolution patches in the high resolution frames are cropped as ground-truth. They set the batch size to 1 and train the model for total 141,360 iterations. For the training of IconVSR, they random extract 15 continuous frame from each super-resolved sequences and crop them to $128\times 128$. The batch size is 1 and iterations is 13,860. During the test phase, they feed 100 consecutive frames into BasicVSR++ to acquire the super-resolved videos. Then, we input them into IconVSR for refining and obtain the final results with rich detailed information.

\subsection{HyperPixel Team}

The HyperPixel team proposes the enhanced BasicVSR++ method, which improves BasicVSR++~\cite{chan2021basicvsr++} from two aspects. On the one hand, to make the network more accurately capture the features of video frames, the Deformable ConvNets v2~\cite{zhu2019deformable} is adopted for spatial feature extraction. On the other hand, in order to intensify the training samples and make their flow more fluently, an intermediate auxiliary $\times 2$ stage loss is added in training. Specifically, they downsample the high quality image in half and use it as the ground-truth in $\times 2$ super resolution in the reconstruction stage of BasicVSR++.

\subsection{CVStars Team}

\begin{figure}[t]
    \centering
    \includegraphics[width=1\linewidth]{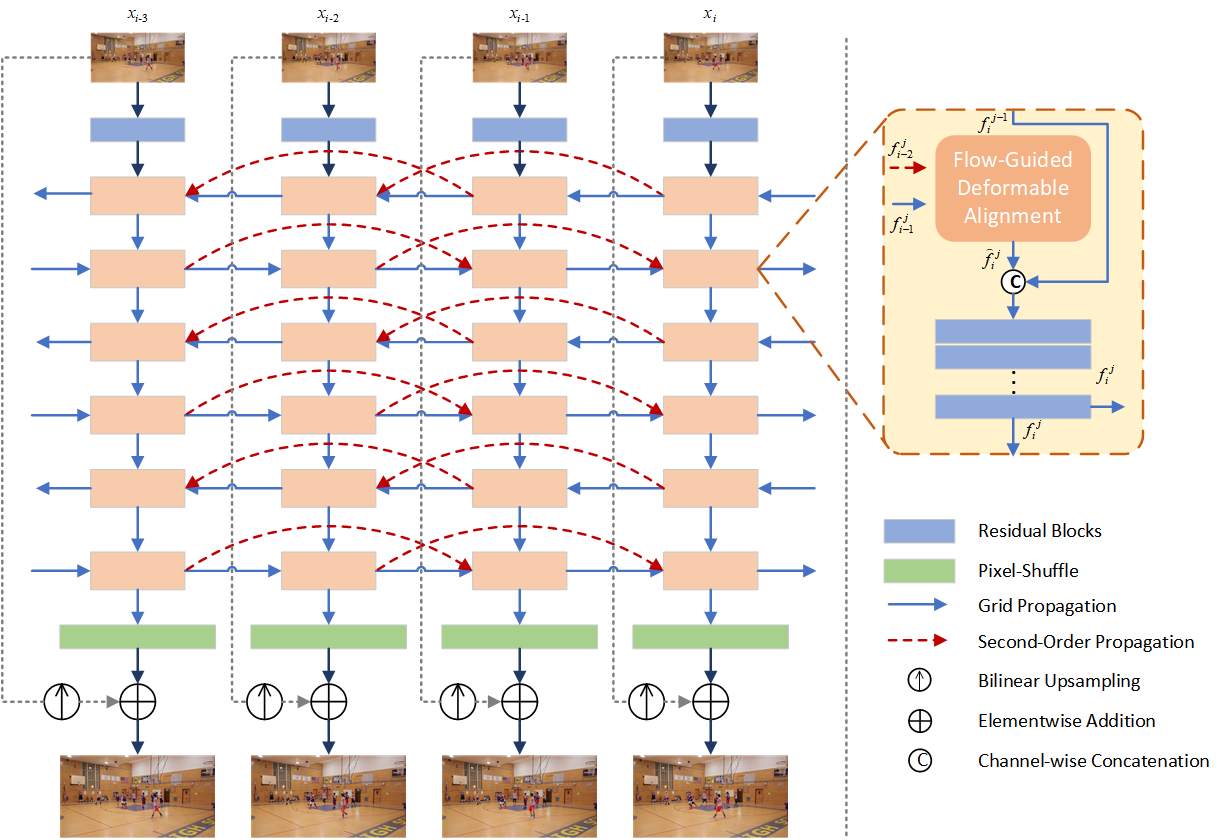}
    \caption{The method proposed by the CVStars Team.}
    \label{fig:cvstars}
\end{figure}

The CVStars Team adopts the BasicVSR++~\cite{chan2021basicvsr++} network and proposes an enhanced version of BasicVSR++. The proposed method is illustrated in Figure~\ref{fig:cvstars}. In order to better extract features and enhance the ability to align between frames, they deepen the depth of forward and back propagation to make them to be triple propagation. In training, they use frequency reconstruction loss to enhance the recovery of high-frequency details. Self-ensemble~\cite{timofte2016seven} and epoch-ensemble methods are used in the testing phase.

\subsection{AVRT Team}

\begin{figure}[t]
    \centering
    \includegraphics[width=\linewidth]{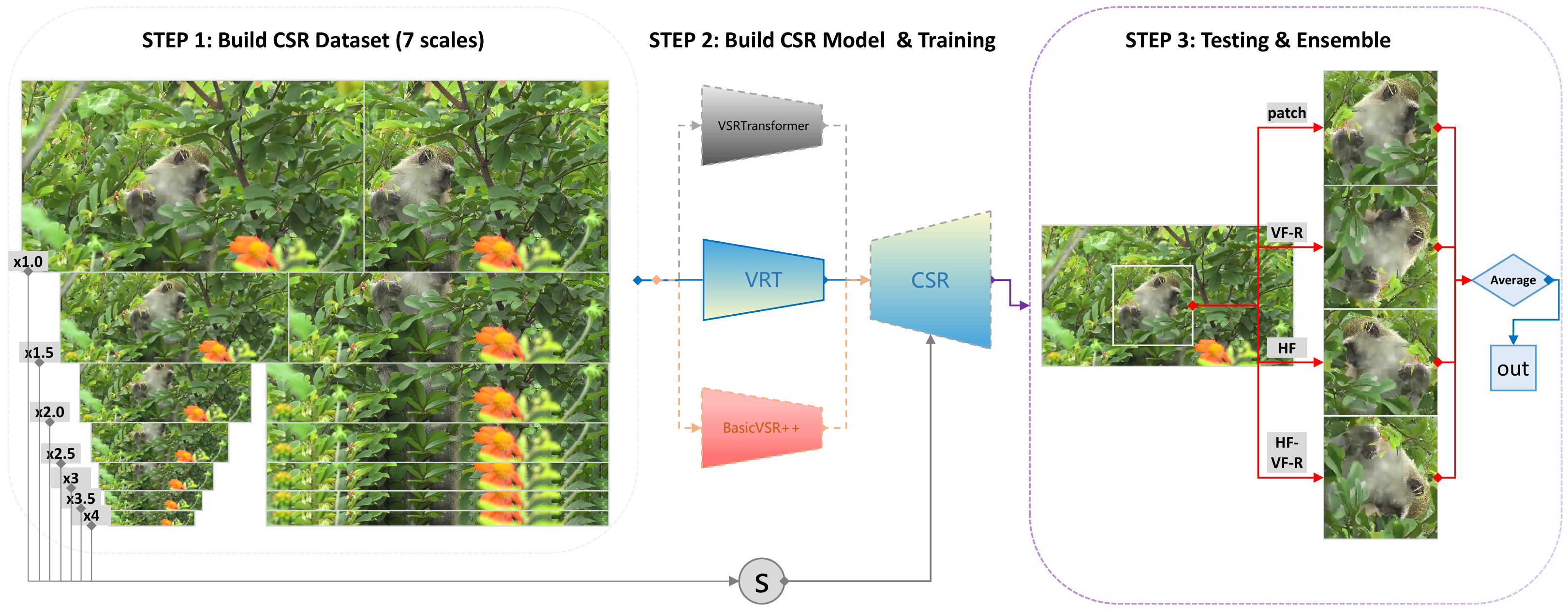}
    \caption{The proposed method of the AVRT Team.}
    \label{fig:AVRT}
\end{figure}

The solution of the AVRT team has three steps as shown in Figure~\ref{fig:AVRT}. First, they build a Continuous Super-Resolution (CSR) dataset with seven scales. Then, they build several CSR models with different feature extraction backbones and train them. Finally, they execute the test using ensemble strategies.

\textbf{Continuous super-resolution datasets}: They  reprocess all the official datasets to generate the CSR datasets by cropping the raw videos in Track 1 to the target resolutions as high quality data and downsample them as the inputs of the encoder to obtain the low resolution and low quality data. Considering the resolution has to be integer numnbers, they simplify the proposed CSR to seven specific super-resolution scales, \ie, 1, 1.5, 2, 2.5, 3, 3.5, 4, which cover the scales of all three tracks. Depending on the official training dataset, they produce four extra pairs of datasets using for 1.5, 2.5, 3 and 3.5 times super resolution tasks whose resolutions are $960\times 528$, $960\times 520$, $960\times 528$ and $952\times 532$, respectively. All these extra datas are processed following the procedure in \cite{yang2021dataset}.

\textbf{Dataset expansion}: In addition to the NTIRE training they collect other videos from Youtube and Movies to expand the scenes in the training dataset. In the first few tests, the PSNR of some categories, such as \textit{Sports} are significantly lower than others, and therefore they drag down the overall performance. Thus, these categories are taken into account when expanding training sets by controlling their quantity to the rest categories as 2:1. All collected videos are cut to clips with 300 to 600 frames by a scene-cut detection algorithm to avoid that a single video clip contains more than one scenes. Then these clips are processed according the above procedure, and finally, 740 groups of extra samples are incorporated into training.

\subsection{StarRay Team}

The StarRay Team employs an enhancement-first strategy based on the enhanced BasicVSR++ model. Specifically, they first use a pretrained STDF~\cite{deng2020spatio} model to enhance the videos and recover the details. Sequentially, they train an enhanced BasicVSR++~\cite{chan2021basicvsr++} model using the Charbonnier loss and the Gradient-Weighted (GW) loss~\cite{wei2020component} based on the enhanced results. To obtain the final results, they ensemble the results of the two models trained by the Charbonnier loss and the GW loss, respectively.

\subsection{Modern\_SR Team}

\begin{figure}[t]
    \centering
    \includegraphics[width=\linewidth]{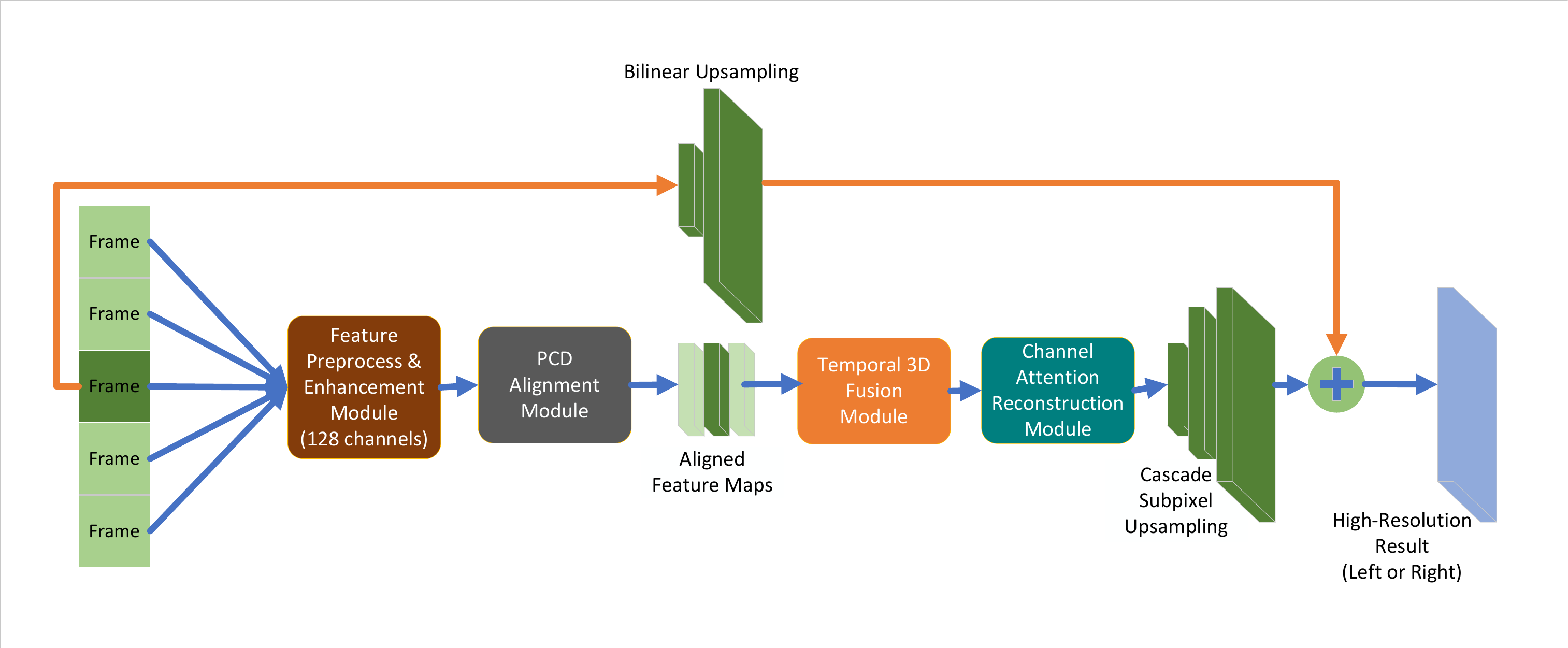}
    \caption{The proposed method of the Modern\_SR Team.}
    \label{fig:modern}
\end{figure}

The Modern\_SR Team designs a method based on the improved EDVR model~\cite{huang2022improved}. In Track 3, they extract a 128-channel feature in super-resolution and take five consecutive video frames as input. The architecture of the proposed method is shown in Figure~\ref{fig:modern}. In comparison with~\cite{huang2022improved}, the proposed method has larger feature size, because the dataset in this challenge is much more complex than the REDS dataset which is used to train~\cite{huang2022improved}.

\subsection{TUK-IKLAB Team}

\begin{figure}[t]
\centering
\includegraphics[width=\linewidth]{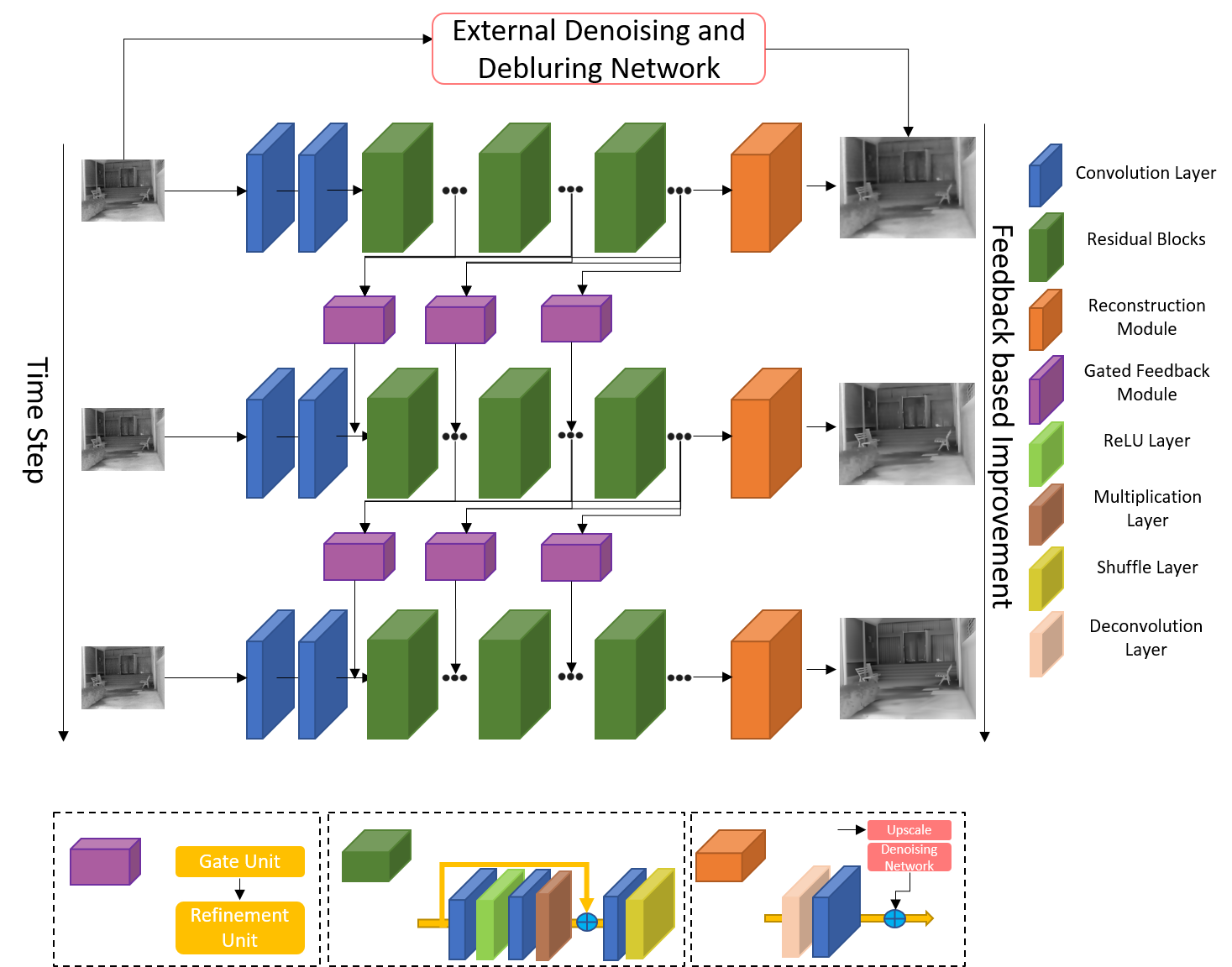}
\caption{Proposed RDFDBK-Net of the TUK-IKLAB Team.}\label{IK}
\end{figure}

The TUK-IKLAB team proposes Residual Denoising and Feedback Networks (RDFDBK-Net) for video super resolution. In short, the method applies denoising, deblurring, enhancement, and spatial-temporal super-resolution on video frames.
The method uses an external denoiser on low-resolution images to remove unwanted noise. As a result, the images are enhanced and scaled. Following this step, the network uses a feedback network using residual frames to improve the enhancement and scaling results. The method is inspired by SRFBN~\cite{li2019feedback} and GMFN~\cite{li2019gated}. The problem with SRFBN was that it only propagates the highest-level features, thus ignoring the low-level features, which results in color enhancement but the preservation of some features such as face or shape is not performed. This also results in the addition of false artifacts. The GMFN on the other hand, added multiple feedback connections to transmit multiple high-level features. Although the method performs slightly better in preserving low-level information, the artifacts are still introduced with more blocking effect. In this regard, they modify the residual blocks in the GMFN to help refine the low-level features while adding an attention mechanism to the refined features. The residual blocks are closely related to the ones used in EDSR network~\cite{lim2017enhanced}. Subsequently, they also use enhancement block to denoise, deblur, and deblock the frames with every feedback step. In addition to the bicubic interpolation-based degradation method, they also add kernel estimation that estimates the blur and noise level for enhancing the LR images. The use of aforementioned methods provide distinctive advantage over SRFBN and GMFN methods, respectively. The proposed network architecture has been shown in Figure~\ref{IK}.

\section*{Acknowledgments}

We thank the NTIRE 2022 sponsors: Huawei, Reality Labs, Bending Spoons, MediaTek, OPPO, Oddity, Voyage81, ETH Zurich (Computer Vision Lab) and University of Wurzburg (CAIDAS).
We also thank Peilin Chen and Prof. Shiqi Wang from the City University of Hong Kong for providing the results of their method~\cite{chen2021compressed} on the validation and test sets.

\

\section*{Appendix: Teams and affiliations}

\subsection*{NTIRE 2022 Team}

\noindent\textit{\textbf{Challenge:}}  

\noindent NTIRE 2022 Challenge on Super-Resolution and Quality Enhancement of Compressed Video

\noindent\textit{\textbf{Organizer(s):}} 

\noindent Ren Yang$^{1}$ ({\texttt{ren.yang@vision.ee.ethz.ch}}), 

\noindent Radu Timofte$^{1,2}$ ({\texttt{radu.timofte@uni-wuerzburg.de}})

\noindent\textit{\textbf{Affiliation(s):}} 

\noindent $^1$ Computer Vision Lab, ETH Z\"urich, Switzerland\\
\noindent $^2$ Julius Maximilian University of W\"urzburg, Germany

\

\subsection*{TaoMC2 Team}
\vspace{-.3em}
\noindent\textit{\textbf{Member(s):}} 

\noindent Meisong Zheng 
\\(\texttt{zhengmeisong.zms@alibaba-inc.com}), \\Qunliang Xing, Minglang Qiao, Mai Xu, Lai Jiang, Huaida Liu, Ying Chen

\noindent\textit{\textbf{Affiliation(s):}} 

\noindent Department of Tao Technology, Alibaba Group; Beihang University, Beijing, China

\
\vspace{-.3em}
\subsection*{GY-Lab Team}
\vspace{-.3em}
\noindent\textit{\textbf{Member(s):}} 

\noindent Chen Fu (\texttt{chenfu@tencent.com}), Youcheng Ben, Xiao Zhou, Pei Cheng, Gang Yu

\noindent\textit{\textbf{Affiliation(s):}} 

\noindent Tencent, China

\
\vspace{-.3em}
\subsection*{HIT\&ACE Team}
\vspace{-.3em}
\noindent\textit{\textbf{Member(s):}} 

\noindent Junyi Li$^1$ (\texttt{nagejacob@gmail.com}), Renlong Wu$^1$, Zhilu Zhang$^1$, Wei Shang$^1$, Zhengyao Lv$^1$, Yunjin Chen, Mingcai Zhou, Dongwei Ren$^1$, Kai Zhang$^2$, Wangmeng Zuo$^1$.

\noindent\textit{\textbf{Affiliation(s):}} 

\noindent $^1$ Harbin Institute of Technology, Harbin, China \\
$^2$ Computer Vision Lab, ETH Zurich, Zurich, Switzerland

\
\vspace{-.3em}
\subsection*{NoahTerminalCV Team}
\vspace{-.3em}
\noindent\textit{\textbf{Member(s):}} 

\noindent Pavel Ostyakov (\texttt{ostyakov.pavel@huawei.com}), Vyal Dmitry, Shakarim Soltanayev, Chervontsev Sergey, Zhussip Magauiya, Xueyi Zou, Youliang Yan

\noindent\textit{\textbf{Affiliation(s):}} 

\noindent Huawei Noah's Ark Lab

\
\vspace{-.3em}
\subsection*{BOE-IOT-AIBD Team}
\vspace{-.3em}
\noindent\textit{\textbf{Member(s):}} 

\noindent Pablo Navarrete Michelini (\texttt{pnavarre@gmail.com}), Yunhua Lu

\noindent\textit{\textbf{Affiliation(s):}} 

\noindent BOE Technology Group Co., Ltd., Beijing, China

\
\vspace{-.3em}
\subsection*{ZX\_VIP Team}
\vspace{-.3em}
\noindent\textit{\textbf{Member(s):}} 

\noindent Diankai Zhang (\texttt{zhang.diankai@zte.com.cn}), Shaoli Liu, Si Gao, Biao Wu, Chengjian Zheng, Xiaofeng Zhang, Kaidi Lu, Ning Wang

\noindent\textit{\textbf{Affiliation(s):}} 

\noindent Audio \& Video Technology Platform Dept., ZTE, Nanjing, China

\subsection*{OCL\_VCE Team}

\noindent\textit{\textbf{Member(s):}} 

\noindent Thuong Nguyen Canh\\ (\texttt{ngcthuong@ids.osaka-u.ac.jp}),\\ Thong Bach

\noindent\textit{\textbf{Affiliation(s):}} 

\noindent Osaka University, Osaka, Japan

\

\subsection*{Trick collector Team}

\noindent\textit{\textbf{Member(s):}} 

\noindent Qing Wang (\texttt{wangqing.keen@bytedance.com}), Xiaopeng Sun, Haoyu Ma, Shijie Zhao, Junlin Li

\noindent\textit{\textbf{Affiliation(s):}} 

\noindent ByteDance, Shenzhen, China

\

\subsection*{XPixel Team}

\noindent\textit{\textbf{Member(s):}} 

\noindent Liangbin Xie (\texttt{lb.xie@siat.ac.cn}), Shuwei Shi, Yujiu Yang, Xintao Wang, Jinjin Gu, Chao Dong

\noindent\textit{\textbf{Affiliation(s):}} 

\noindent Shanghai AI Lab, Shanghai, China; Shenzhen Institutes of Advanced Technology (SIAT), Chinese Academy of Sciences (CAS), Shenzhen, China

\

\subsection*{OREO Team}

\noindent\textit{\textbf{Member(s):}} 

\noindent Xiaodi Shi (\texttt{sxd0071@gmail.com}), Chunmei Nian, Dong Jiang, Jucai Lin

\noindent\textit{\textbf{Affiliation(s):}} 

\noindent Hangzhou University of Electronic Science and technology, Hangzhou, China

\

\subsection*{HMSR Team}

\noindent\textit{\textbf{Member(s):}} 

\noindent Zhihuai Xie (\texttt{xzhuai520@163.com})

\

\subsection*{UESTC+XJU CV Team}
\vspace{-.3em}
\noindent\textit{\textbf{Member(s):}} 

\noindent Mao Ye$^1$ (\texttt{cvlab.uestc@gmail.com}), Dengyan Luo$^1$, Shengjie Chen$^1$, Liuhan Peng$^2$

\noindent\textit{\textbf{Affiliation(s):}} 

\noindent $^1$ University of Electronic Science and Technology of China, Chengdu, China\\
$^2$ Xinjiang University, Xinjiang, China.

\

\subsection*{TBE Team}
\vspace{-.3em}
\noindent\textit{\textbf{Member(s):}} 

\noindent Xin Liu (\texttt{liuxin9976@163.com})

\

\subsection*{HyperPixel Team}
\vspace{-.3em}
\noindent\textit{\textbf{Member(s):}} 

\noindent Qian Wang (\texttt{utopiamail@qq.com}), Xin Liu

\noindent\textit{\textbf{Affiliation(s):}} 

\noindent China Mobile Research Institute

\

\subsection*{CVStars Team}
\vspace{-.3em}
\noindent\textit{\textbf{Member(s):}} 

\noindent Boyang Liang (\texttt{951240992@qq.com}), Dong Hang, Yuhao Huang, Kai Chen



\

\subsection*{StarRay Team}
\vspace{-.3em}
\noindent\textit{\textbf{Member(s):}} 

\noindent Xingbei Guo (\texttt{guoxb7@mail2.sysu.edu.cn}), Yujing Sun, Huilei Wu, Pengxu Wei

\noindent\textit{\textbf{Affiliation(s):}} 

\noindent Sun Yat-sen University, Guangzhou, China; Peking University, Beijing, China

\

\subsection*{Modern\_SR Team}
\vspace{-.3em}
\noindent\textit{\textbf{Member(s):}} 

\noindent Yulin Huang (\texttt{815018345@qq.com}), Junying Chen

\noindent\textit{\textbf{Affiliation(s):}} 

\noindent City University of Hong Kong, Hong Kong SAR, China; South China University of Technology, Guangzhou, China

\

\subsection*{TUK-IKLAB Team}
\vspace{-.3em}
\noindent\textit{\textbf{Member(s):}} 

\noindent Ik Hyun Lee$^{1,3}$  (\texttt{ihlee@tukorea.ac.kr}), Sunder Ali Khowaja$^{1,2}$, Jiseok Yoon$^1$ 

\noindent\textit{\textbf{Affiliation(s):}} 

\noindent $^1$ Tech University of Korea, Siheung-si, Korea \\ $^2$ University of Sindh, Pakistan \\$^3$ IKLAB

\

{\small
\bibliographystyle{ieee_fullname}
\bibliography{egbib}

\begin{thebibliography}{10}\itemsep=-1pt

\bibitem{arad2022ntiredemosaicing}
Boaz Arad, Radu Timofte, Rony Yahel, Nimrod Morag, Amir Bernat, et~al.
\newblock {NTIRE} 2022 spectral demosaicing challenge and dataset.
\newblock In {\em Proceedings of the IEEE/CVF Conference on Computer Vision and
  Pattern Recognition (CVPR) Workshops}, 2022.

\bibitem{arad2022ntirerecovery}
Boaz Arad, Radu Timofte, Rony Yahel, Nimrod Morag, Amir Bernat, et~al.
\newblock {NTIRE} 2022 spectral recovery challenge and dataset.
\newblock In {\em Proceedings of the IEEE/CVF Conference on Computer Vision and
  Pattern Recognition (CVPR) Workshops}, 2022.

\bibitem{ba2016layer}
Jimmy~Lei Ba, Jamie~Ryan Kiros, and Geoffrey~E Hinton.
\newblock Layer normalization.
\newblock {\em arXiv preprint arXiv:1607.06450}, 2016.

\bibitem{bhat2021ntire}
Goutam Bhat, Martin Danelljan, and Radu Timofte.
\newblock {NTIRE} 2021 challenge on burst super-resolution: Methods and
  results.
\newblock In {\em Proceedings of the IEEE/CVF Conference on Computer Vision and
  Pattern Recognition (CVPR) Workshops}, pages 613--626, 2021.

\bibitem{bhat2022ntire}
Goutam Bhat, Martin Danelljan, Radu Timofte, et~al.
\newblock {NTIRE} 2022 burst super-resolution challenge.
\newblock In {\em Proceedings of the IEEE/CVF Conference on Computer Vision and
  Pattern Recognition (CVPR) Workshops}, 2022.

\bibitem{DBLP:conf/sisap/BoytsovN13}
Leonid Boytsov and Bilegsaikhan Naidan.
\newblock Engineering efficient and effective non-metric space library.
\newblock In Nieves~R. Brisaboa, Oscar Pedreira, and Pavel Zezula, editors,
  {\em Similarity Search and Applications - 6th International Conference,
  {SISAP} 2013, {A} Coru{\~{n}}a, Spain, October 2-4, 2013, Proceedings},
  volume 8199 of {\em Lecture Notes in Computer Science}, pages 280--293.
  Springer, 2013.

\bibitem{briechle2001template}
Kai Briechle and Uwe~D Hanebeck.
\newblock Template matching using fast normalized cross correlation.
\newblock In {\em Optical Pattern Recognition XII}, volume 4387, pages 95--102.
  International Society for Optics and Photonics, 2001.

\bibitem{caballero2017real}
Jose Caballero, Christian Ledig, Andrew Aitken, Alejandro Acosta, Johannes
  Totz, Zehan Wang, and Wenzhe Shi.
\newblock Real-time video super-resolution with spatio-temporal networks and
  motion compensation.
\newblock In {\em Proceedings of the IEEE Conference on Computer Vision and
  Pattern Recognition (CVPR)}, pages 4778--4787, 2017.

\bibitem{chan2021basicvsr}
Kelvin~CK Chan, Xintao Wang, Ke Yu, Chao Dong, and Chen~Change Loy.
\newblock Basic{VSR}: The search for essential components in video
  super-resolution and beyond.
\newblock In {\em Proceedings of the IEEE/CVF Conference on Computer Vision and
  Pattern Recognition (CVPR)}, 2021.

\bibitem{chan2021basicvsr++}
Kelvin~CK Chan, Shangchen Zhou, Xiangyu Xu, and Chen~Change Loy.
\newblock Basicvsr++: Improving video super-resolution with enhanced
  propagation and alignment.
\newblock {\em arXiv preprint arXiv:2104.13371}, 2021.

\bibitem{charbonnier1994two}
Pierre Charbonnier, Laure Blanc-Feraud, Gilles Aubert, and Michel Barlaud.
\newblock Two deterministic half-quadratic regularization algorithms for
  computed imaging.
\newblock In {\em Proceedings of 1st International Conference on Image
  Processing (ICIP)}, volume~2, pages 168--172. IEEE, 1994.

\bibitem{chen2020bitstream}
Peilin Chen, Wenhan Yang, Long Sun, and Shiqi Wang.
\newblock When bitstream prior meets deep prior: Compressed video
  super-resolution with learning from decoding.
\newblock In {\em Proceedings of the ACM International Conference on Multimedia
  (ACM MM)}, pages 1000--1008, 2020.

\bibitem{chen2021compressed}
Peilin Chen, Wenhan Yang, Meng Wang, Long Sun, Kangkang Hu, and Shiqi Wang.
\newblock Compressed domain deep video super-resolution.
\newblock {\em IEEE Transactions on Image Processing}, 30:7156--7169, 2021.

\bibitem{chen2020simple}
Ting Chen, Simon Kornblith, Mohammad Norouzi, and Geoffrey Hinton.
\newblock A simple framework for contrastive learning of visual
  representations.
\newblock In {\em International Conference on Machine Learning (ICML)}, pages
  1597--1607. PMLR, 2020.

\bibitem{dai2017deformable}
Jifeng Dai, Haozhi Qi, Yuwen Xiong, Yi Li, Guodong Zhang, Han Hu, and Yichen
  Wei.
\newblock Deformable convolutional networks.
\newblock In {\em Proceedings of the IEEE International Conference on Computer
  Vision (ICCV)}, pages 764--773, 2017.

\bibitem{deng2020spatio}
Jianing Deng, Li Wang, Shiliang Pu, and Cheng Zhuo.
\newblock Spatio-temporal deformable convolution for compressed video quality
  enhancement.
\newblock {\em Proceedings of the AAAI Conference on Artificial Intelligence},
  34(07):10696--10703, 2020.

\bibitem{elfwing2018sigmoid}
Stefan Elfwing, Eiji Uchibe, and Kenji Doya.
\newblock Sigmoid-weighted linear units for neural network function
  approximation in reinforcement learning.
\newblock {\em Neural Networks}, 107:3--11, 2018.

\bibitem{ershov2022ntire}
Egor Ershov, Alex Savchik, Denis Shepelev, Nikola Banic, Michael~S Brown, Radu
  Timofte, et~al.
\newblock {NTIRE} 2022 challenge on night photography rendering.
\newblock In {\em Proceedings of the IEEE/CVF Conference on Computer Vision and
  Pattern Recognition (CVPR) Workshops}, 2022.

\bibitem{gu2022ntire}
Jinjin Gu, Haoming Cai, Chao Dong, Jimmy Ren, Radu Timofte, et~al.
\newblock {NTIRE} 2022 challenge on perceptual image quality assessment.
\newblock In {\em Proceedings of the IEEE/CVF Conference on Computer Vision and
  Pattern Recognition (CVPR) Workshops}, 2022.

\bibitem{guan2019mfqe}
Zhenyu Guan, Qunliang Xing, Mai Xu, Ren Yang, Tie Liu, and Zulin Wang.
\newblock {MFQE} 2.0: A new approach for multi-frame quality enhancement on
  compressed video.
\newblock {\em IEEE Transactions on Pattern Analysis and Machine Intelligence},
  2019.

\bibitem{haris2019recurrent}
Muhammad Haris, Gregory Shakhnarovich, and Norimichi Ukita.
\newblock Recurrent back-projection network for video super-resolution.
\newblock In {\em Proceedings of the IEEE/CVF Conference on Computer Vision and
  Pattern Recognition (CVPR)}, pages 3897--3906, 2019.

\bibitem{he2015delving}
Kaiming He, Xiangyu Zhang, Shaoqing Ren, and Jian Sun.
\newblock Delving deep into rectifiers: Surpassing human-level performance on
  imagenet classification.
\newblock In {\em Proceedings of the IEEE International Conference on Computer
  Vision (ICCV)}, pages 1026--1034, 2015.

\bibitem{he2016deep}
Kaiming He, Xiangyu Zhang, Shaoqing Ren, and Jian Sun.
\newblock Deep residual learning for image recognition.
\newblock In {\em Proceedings of the IEEE Conference on Computer Vision and
  Pattern Recognition (CVPR)}, pages 770--778, 2016.

\bibitem{huang2017arbitrary}
Xun Huang and Serge Belongie.
\newblock Arbitrary style transfer in real-time with adaptive instance
  normalization.
\newblock In {\em Proceedings of the IEEE International Conference on Computer
  Vision (ICCV)}, pages 1501--1510, 2017.

\bibitem{huang2022improved}
Yulin Huang and Junying Chen.
\newblock Improved {EDVR} model for robust and efficient video
  super-resolution.
\newblock In {\em Proceedings of the IEEE/CVF Winter Conference on Applications
  of Computer Vision (WACV) Workshops)}, pages 103--111, 2022.

\bibitem{huang2015bidirectional}
Yan Huang, Wei Wang, and Liang Wang.
\newblock Bidirectional recurrent convolutional networks for multi-frame
  super-resolution.
\newblock {\em Advances in Neural Information Processing Systems (NeurIPS)},
  28, 2015.

\bibitem{huo2021recurrent}
Yongkai Huo, Qiyan Lian, Shaoshi Yang, and Jianmin Jiang.
\newblock A recurrent video quality enhancement framework with
  multi-granularity frame-fusion and frame difference based attention.
\newblock {\em Neurocomputing}, 431:34--46, 2021.

\bibitem{youtube}
Google Inc.
\newblock {Y}ou{T}ube.
\newblock \url{https://www.youtube.com}.

\bibitem{jiang2021learning}
Shihao Jiang, Dylan Campbell, Yao Lu, Hongdong Li, and Richard Hartley.
\newblock Learning to estimate hidden motions with global motion aggregation.
\newblock In {\em Proceedings of the IEEE/CVF International Conference on
  Computer Vision (ICCV)}, pages 9772--9781, 2021.

\bibitem{jo2018deep}
Younghyun Jo, Seoung~Wug Oh, Jaeyeon Kang, and Seon~Joo Kim.
\newblock Deep video super-resolution network using dynamic upsampling filters
  without explicit motion compensation.
\newblock In {\em Proceedings of the IEEE Conference on Computer Vision and
  Pattern Recognition (CVPR)}, pages 3224--3232, 2018.

\bibitem{kingma2014adam}
Diederik~P Kingma and Jimmy Ba.
\newblock Adam: A method for stochastic optimization.
\newblock In {\em Proceedings of the International Conference on Learning
  Representations (ICLR)}, 2015.

\bibitem{kong2021reflash}
Xiangtao Kong, Xina Liu, Jinjin Gu, Yu Qiao, and Chao Dong.
\newblock Reflash dropout in image super-resolution.
\newblock {\em arXiv preprint arXiv:2112.12089}, 2021.

\bibitem{lai2018fast}
Wei-Sheng Lai, Jia-Bin Huang, Narendra Ahuja, and Ming-Hsuan Yang.
\newblock Fast and accurate image super-resolution with deep laplacian pyramid
  networks.
\newblock {\em IEEE Transactions on Pattern Analysis and Machine Intelligence},
  41(11):2599--2613, 2018.

\bibitem{lample2019large}
Guillaume Lample, Alexandre Sablayrolles, Marc'Aurelio Ranzato, Ludovic
  Denoyer, and Herv{\'e} J{\'e}gou.
\newblock Large memory layers with product keys.
\newblock {\em Advances in Neural Information Processing Systems (NeurIPS)},
  32, 2019.

\bibitem{li2019gated}
Qilei Li, Zhen Li, Lu Lu, Gwanggil Jeon, Kai Liu, and Xiaomin Yang.
\newblock Gated multiple feedback network for image super-resolution.
\newblock {\em arXiv preprint arXiv:1907.04253}, 2019.

\bibitem{li2022ntire}
Yawei Li, Kai Zhang, Radu Timofte, Luc Van~Gool, et~al.
\newblock {NTIRE} 2022 challenge on efficient super-resolution: Methods and
  results.
\newblock In {\em Proceedings of the IEEE/CVF Conference on Computer Vision and
  Pattern Recognition (CVPR) Workshops}, 2022.

\bibitem{li2019feedback}
Zhen Li, Jinglei Yang, Zheng Liu, Xiaomin Yang, Gwanggil Jeon, and Wei Wu.
\newblock Feedback network for image super-resolution.
\newblock In {\em Proceedings of the IEEE/CVF Conference on Computer Vision and
  Pattern Recognition (CVPR)}, pages 3867--3876, 2019.

\bibitem{liang2021swinir}
Jingyun Liang, Jiezhang Cao, Guolei Sun, Kai Zhang, Luc Van~Gool, and Radu
  Timofte.
\newblock Swinir: Image restoration using swin transformer.
\newblock In {\em Proceedings of the IEEE/CVF International Conference on
  Computer Vision (ICCV)}, pages 1833--1844, 2021.

\bibitem{lim2017enhanced}
Bee Lim, Sanghyun Son, Heewon Kim, Seungjun Nah, and Kyoung Mu~Lee.
\newblock Enhanced deep residual networks for single image super-resolution.
\newblock In {\em Proceedings of the IEEE Conference on Computer Vision and
  Pattern Recognition (CVPR) Workshops}, pages 136--144, 2017.

\bibitem{lin2022revisiting}
Zudi Lin, Prateek Garg, Atmadeep Banerjee, Salma~Abdel Magid, Deqing Sun, Yulun
  Zhang, Luc Van~Gool, Donglai Wei, and Hanspeter Pfister.
\newblock Revisiting rcan: Improved training for image super-resolution.
\newblock {\em arXiv preprint arXiv:2201.11279}, 2022.

\bibitem{RevRCAN}
Zudi Lin, Prateek Garg, Atmadeep Banerjee, Salma~Abdel Magid, Deqing Sun, Yulun
  Zhang, Luc Van~Gool, Donglai Wei, and Hanspeter Pfister.
\newblock Revisiting rcan: Improved training for image super-resolution.
\newblock {\em arXiv preprint arXiv:2201.11279}, 2022.

\bibitem{liu2017robust}
Ding Liu, Zhaowen Wang, Yuchen Fan, Xianming Liu, Zhangyang Wang, Shiyu Chang,
  and Thomas Huang.
\newblock Robust video super-resolution with learned temporal dynamics.
\newblock In {\em Proceedings of the IEEE International Conference on Computer
  Vision (ICCV)}, pages 2507--2515, 2017.

\bibitem{ConvNEXT}
Zhuang Liu, Hanzi Mao, Chao-Yuan Wu, Christoph Feichtenhofer, Trevor Darrell,
  and Saining Xie.
\newblock A convnet for the 2020s.
\newblock {\em arXiv preprint arXiv:2201.03545}, 2022.

\bibitem{liu2021video}
Ze Liu, Jia Ning, Yue Cao, Yixuan Wei, Zheng Zhang, Stephen Lin, and Han Hu.
\newblock Video swin transformer.
\newblock {\em arXiv preprint arXiv:2106.13230}, 2021.

\bibitem{loshchilov2016sgdr}
Ilya Loshchilov and Frank Hutter.
\newblock {SGDR}: Stochastic gradient descent with warm restarts.
\newblock {\em arXiv preprint arXiv:1608.03983}, 2016.

\bibitem{lu2018deep}
Guo Lu, Wanli Ouyang, Dong Xu, Xiaoyun Zhang, Zhiyong Gao, and Ming-Ting Sun.
\newblock Deep {Kalman} filtering network for video compression artifact
  reduction.
\newblock In {\em Proceedings of the European Conference on Computer Vision
  (ECCV)}, pages 568--584, 2018.

\bibitem{lugmayr2022ntire}
Andreas Lugmayr, Martin Danelljan, Radu Timofte, et~al.
\newblock {NTIRE} 2022 challenge on learning the super-resolution space.
\newblock In {\em Proceedings of the IEEE/CVF Conference on Computer Vision and
  Pattern Recognition (CVPR) Workshops}, 2022.

\bibitem{DBLP:journals/corr/MalkovY16}
Yury~A. Malkov and Dmitry~A. Yashunin.
\newblock Efficient and robust approximate nearest neighbor search using
  hierarchical navigable small world graphs.
\newblock {\em CoRR}, abs/1603.09320, 2016.

\bibitem{nah2019ntire}
Seungjun Nah, Sungyong Baik, Seokil Hong, Gyeongsik Moon, Sanghyun Son, Radu
  Timofte, and Kyoung Mu~Lee.
\newblock {NTIRE} 2019 challenge on video deblurring and super-resolution:
  Dataset and study.
\newblock In {\em Proceedings of the IEEE/CVF Conference on Computer Vision and
  Pattern Recognition (CVPR) Workshops}, pages 0--0, 2019.

\bibitem{perezpellitero2022ntire}
Eduardo P\'erez-Pellitero, Sibi Catley-Chandar, Richard Shaw, Ales Leonardis,
  Radu Timofte, et~al.
\newblock {NTIRE} 2022 challenge on high dynamic range imaging: Methods and
  results.
\newblock In {\em Proceedings of the IEEE/CVF Conference on Computer Vision and
  Pattern Recognition (CVPR) Workshops}, 2022.

\bibitem{ranjan2017optical}
Anurag Ranjan and Michael~J Black.
\newblock Optical flow estimation using a spatial pyramid network.
\newblock In {\em Proceedings of the IEEE Conference on Computer Vision and
  Pattern Recognition (CVPR)}, pages 4161--4170, 2017.

\bibitem{rao2020bitstream}
Rakesh Rao~Ramachandra Rao, Steve G{\"o}ring, Peter List, Werner Robitza,
  Bernhard Feiten, Ulf W{\"u}stenhagen, and Alexander Raake.
\newblock Bitstream-based model standard for 4k/uhd: Itu-t p. 1204.3—model
  details, evaluation, analysis and open source implementation.
\newblock In {\em 2020 Twelfth International Conference on Quality of
  Multimedia Experience (QoMEX)}, pages 1--6. IEEE, 2020.

\bibitem{romero2022ntire}
Andres Romero, Angela Castillo, Jose~M Abril-Nova, Radu Timofte, et~al.
\newblock {NTIRE} 2022 image inpainting challenge: Report.
\newblock In {\em Proceedings of the IEEE/CVF Conference on Computer Vision and
  Pattern Recognition (CVPR) Workshops}, 2022.

\bibitem{sajjadi2018frame}
Mehdi~SM Sajjadi, Raviteja Vemulapalli, and Matthew Brown.
\newblock Frame-recurrent video super-resolution.
\newblock In {\em Proceedings of the IEEE Conference on Computer Vision and
  Pattern Recognition (CVPR)}, pages 6626--6634, 2018.

\bibitem{tao2017detail}
Xin Tao, Hongyun Gao, Renjie Liao, Jue Wang, and Jiaya Jia.
\newblock Detail-revealing deep video super-resolution.
\newblock In {\em Proceedings of the IEEE International Conference on Computer
  Vision (ICCV)}, pages 4472--4480, 2017.

\bibitem{tian2020tdan}
Yapeng Tian, Yulun Zhang, Yun Fu, and Chenliang Xu.
\newblock {TDAN}: Temporally-deformable alignment network for video
  super-resolution.
\newblock In {\em Proceedings of the IEEE/CVF Conference on Computer Vision and
  Pattern Recognition (CVPR)}, pages 3360--3369, 2020.

\bibitem{timofte2016seven}
Radu Timofte, Rasmus Rothe, and Luc Van~Gool.
\newblock Seven ways to improve example-based single image super resolution.
\newblock In {\em Proceedings of the IEEE Conference on Computer Vision and
  Pattern Recognition (CVPR)}, pages 1865--1873, 2016.

\bibitem{tompson2015efficient}
Jonathan Tompson, Ross Goroshin, Arjun Jain, Yann LeCun, and Christoph Bregler.
\newblock Efficient object localization using convolutional networks.
\newblock In {\em Proceedings of the IEEE Conference on Computer Vision and
  Pattern Recognition (CVPR)}, pages 648--656, 2015.

\bibitem{turkowski1990filters}
Ken Turkowski.
\newblock Filters for common resampling tasks.
\newblock {\em Graphics Gems I}, pages 147--165, 1990.

\bibitem{wang2020multi}
Jianyi Wang, Xin Deng, Mai Xu, Congyong Chen, and Yuhang Song.
\newblock Multi-level wavelet-based generative adversarial network for
  perceptual quality enhancement of compressed video.
\newblock In {\em Proceedings of the European Conference on Computer Vision
  (ECCV)}, pages 405--421. Springer, 2020.

\bibitem{wang2020deep}
Longguang Wang, Yulan Guo, Li Liu, Zaiping Lin, Xinpu Deng, and Wei An.
\newblock Deep video super-resolution using hr optical flow estimation.
\newblock {\em IEEE Transactions on Image Processing}, 29:4323--4336, 2020.

\bibitem{wang2022ntire}
Longguang Wang, Yulan Guo, Yingqian Wang, Juncheng Li, Shuhang Gu, Radu
  Timofte, et~al.
\newblock {NTIRE} 2022 challenge on stereo image super-resolution: Methods and
  results.
\newblock In {\em Proceedings of the IEEE/CVF Conference on Computer Vision and
  Pattern Recognition (CVPR) Workshops}, 2022.

\bibitem{wang2017novel}
Tingting Wang, Mingjin Chen, and Hongyang Chao.
\newblock A novel deep learning-based method of improving coding efficiency
  from the decoder-end for {HEVC}.
\newblock In {\em Proceedings of the Data Compression Conference (DCC)}, pages
  410--419. IEEE, 2017.

\bibitem{wang2019edvr}
Xintao Wang, Kelvin~CK Chan, Ke Yu, Chao Dong, and Chen Change~Loy.
\newblock {EDVR}: Video restoration with enhanced deformable convolutional
  networks.
\newblock In {\em Proceedings of the IEEE/CVF Conference on Computer Vision and
  Pattern Recognition (CVPR) Workshops}, 2019.

\bibitem{wei2020component}
Pengxu Wei, Ziwei Xie, Hannan Lu, Zongyuan Zhan, Qixiang Ye, Wangmeng Zuo, and
  Liang Lin.
\newblock Component divide-and-conquer for real-world image super-resolution.
\newblock In {\em European Conference on Computer Vision (ECCV)}, pages
  101--117. Springer, 2020.

\bibitem{xing2020early}
Qunliang Xing, Mai Xu, Tianyi Li, and Zhenyu Guan.
\newblock Early exit or not: resource-efficient blind quality enhancement for
  compressed images.
\newblock In {\em Proceedings of the European Conference on Computer Vision
  (ECCV)}, pages 275--292. Springer, 2020.

\bibitem{xu2021multi}
Mai Xu, Ren Yang, Tie Liu, Tianyi Li, and Zhaoji Fang.
\newblock Multi-frame quality enhancement for compressed video, Mar.~30 2021.
\newblock US Patent 10,965,959.

\bibitem{Xu_2019_ICCV}
Yi Xu, Longwen Gao, Kai Tian, Shuigeng Zhou, and Huyang Sun.
\newblock Non-local {C}onv{LSTM} for video compression artifact reduction.
\newblock In {\em Proceedings of The IEEE International Conference on Computer
  Vision (ICCV)}, October 2019.

\bibitem{xue2019video}
Tianfan Xue, Baian Chen, Jiajun Wu, Donglai Wei, and William~T Freeman.
\newblock Video enhancement with task-oriented flow.
\newblock {\em International Journal of Computer Vision}, 127(8):1106--1125,
  2019.

\bibitem{yang2020learning}
Ren Yang, Fabian Mentzer, Luc~Van Gool, and Radu Timofte.
\newblock Learning for video compression with hierarchical quality and
  recurrent enhancement.
\newblock In {\em Proceedings of the IEEE/CVF Conference on Computer Vision and
  Pattern Recognition (CVPR)}, pages 6628--6637, 2020.

\bibitem{yang2019quality}
Ren Yang, Xiaoyan Sun, Mai Xu, and Wenjun Zeng.
\newblock Quality-gated convolutional {LSTM} for enhancing compressed video.
\newblock In {\em Proceedings of the IEEE International Conference on
  Multimedia and Expo (ICME)}, pages 532--537. IEEE, 2019.

\bibitem{yang2021dataset}
Ren Yang, Radu Timofte, et~al.
\newblock {NTIRE 2021} challenge on quality enhancement of compressed video:
  Dataset and study.
\newblock In {\em Proceedings of the IEEE/CVF Conference on Computer Vision and
  Pattern Recognition (CVPR) Workshops}, 2021.

\bibitem{yang2021ntire}
Ren Yang, Radu Timofte, et~al.
\newblock {NTIRE 2021} challenge on quality enhancement of compressed video:
  Methods and results.
\newblock In {\em IEEE/CVF Conference on Computer Vision and Pattern
  Recognition (CVPR) Workshops}, 2021.

\bibitem{yang2022ntire}
Ren Yang, Radu Timofte, et~al.
\newblock {NTIRE} 2022 challenge on super-resolution and quality enhancement of
  compressed video: Dataset, methods and results.
\newblock In {\em Proceedings of the IEEE/CVF Conference on Computer Vision and
  Pattern Recognition (CVPR) Workshops}, 2022.

\bibitem{yang2018enhancing}
Ren Yang, Mai Xu, Tie Liu, Zulin Wang, and Zhenyu Guan.
\newblock Enhancing quality for {HEVC} compressed videos.
\newblock {\em IEEE Transactions on Circuits and Systems for Video Technology},
  2018.

\bibitem{yang2017decoder}
Ren Yang, Mai Xu, and Zulin Wang.
\newblock Decoder-side {HEVC} quality enhancement with scalable convolutional
  neural network.
\newblock In {\em Proceedings of the IEEE International Conference on
  Multimedia and Expo (ICME)}, pages 817--822. IEEE, 2017.

\bibitem{yang2018multi}
Ren Yang, Mai Xu, Zulin Wang, and Tianyi Li.
\newblock Multi-frame quality enhancement for compressed video.
\newblock In {\em Proceedings of the IEEE Conference on Computer Vision and
  Pattern Recognition (CVPR)}, pages 6664--6673, 2018.

\bibitem{zamir2021restormer}
Syed~Waqas Zamir, Aditya Arora, Salman Khan, Munawar Hayat, Fahad~Shahbaz Khan,
  and Ming-Hsuan Yang.
\newblock Restormer: Efficient transformer for high-resolution image
  restoration.
\newblock {\em arXiv preprint arXiv:2111.09881}, 2021.

\bibitem{zhu2019deformable}
Xizhou Zhu, Han Hu, Stephen Lin, and Jifeng Dai.
\newblock Deformable convnets v2: More deformable, better results.
\newblock In {\em Proceedings of the IEEE/CVF Conference on Computer Vision and
  Pattern Recognition (CVPR)}, pages 9308--9316, 2019.

\end{thebibliography}
}

\end{document}